\documentclass[11pt]{article}

\usepackage[preprint]{acl}

\usepackage{times}
\usepackage{latexsym}
\usepackage{amsmath}
\usepackage{amssymb}
\usepackage{amsfonts}
\usepackage{nicefrac}

\usepackage[T1]{fontenc}
\usepackage[utf8]{inputenc}

\usepackage{microtype}
\usepackage{inconsolata}

\usepackage{graphicx}
\usepackage{xspace}
\usepackage{makecell}
\usepackage{array}
\usepackage{colortbl}
\usepackage{multirow}
\usepackage{subcaption}
\usepackage{tabularx}
\usepackage[dvipsnames]{xcolor}
\usepackage{color}
\usepackage{pifont}
\usepackage{CJKutf8}
\usepackage{enumitem}
\usepackage{diagbox}
\usepackage[flushleft]{threeparttable}
\usepackage{booktabs}
\usepackage{wrapfig}
\usepackage[normalem]{ulem}
\usepackage{dsfont}
\usepackage{marvosym}
\usepackage{algorithmic,algorithm}

\newcommand{\ourmethod}{{\fontfamily{lmtt}\selectfont \textbf{DMoA}}\xspace}
\newcommand{\llmname}[1]{{\texttt{#1}}}

\definecolor{fbApp}{HTML}{ffe4e3}
\definecolor{mydarkblue}{rgb}{0,0.3,0.9}

\newcommand{\rowcr}{\rowcolor{fbApp}}

\newcommand{\rowcb}{\rowcolor{CadetBlue!10}}
\newcommand{\rowcg}{\rowcolor{gray!10}}

\newcommand{\first}[1]{\textcolor{red}{\textbf{#1}}}
\newcommand{\second}[1]{\textcolor{blue}{\underline{#1}}}

\newcommand{\blue}[1]{$_{\color{BlueGreen}\downarrow #1}$}
\newcommand{\red}[1]{$_{\color{RedOrange}\uparrow #1}$}

\usepackage{tikz}
\usepackage{tcolorbox}

\title{Differentiable Mixture-of-Agents Incentivizes Swarm Intelligence \\ of Large Language Models}

\author{
  Xingjian Wu, Junkai Lu, Siyu Yan, Xiangfei Qiu, Jilin Hu, \\ \textbf{Chenjuan Guo}, \textbf{Bin Yang}\textsuperscript{\Letter}
\\
 East China Normal University
\\
   \{xjwu, jklu, syyan, xfqiu\}@stu.ecnu.edu.cn, \\ 
   \{jlhu, cjguo, byang\}@dase.ecnu.edu.cn
}

\begin{document}
\maketitle

\begin{abstract}
Recent advances in Large Language Models (LLMs) have catalyzed the development of multi-agent systems (MAS) for complex reasoning tasks. However, existing MAS typically rely on pre-defined or pre-compiled communication topologies, which limits their flexibility and adaptability to dynamic task requirements. In this work, we propose Differentiable Mixture-of-Agents (\ourmethod), a self-evolving multi-agent framework that enables elastic and adaptive agent collaboration during inference. Instead of statically constructing workflows, \ourmethod dynamically routes and activates agents at each reasoning step, allowing the system to implicitly simulate diverse communication topologies and adapt to evolving demands. To achieve this, we design a differentiable, context-aware routing mechanism that leverages recurrent structures to incorporate historical and contextual information, producing sparse agent activations in a step-wise manner. Furthermore, we introduce predictive entropy as self-supervised signals to optimize the routing process, enabling efficient test-time adaptation without external annotations. Extensive experiments across 9 benchmarks demonstrate that \ourmethod achieves state-of-the-art performance while exhibiting strong efficiency, robustness, and ensembling capabilities.
\end{abstract}

\section{Introduction}
\begin{figure*}[t]
    \centering
\includegraphics[width=1\linewidth]{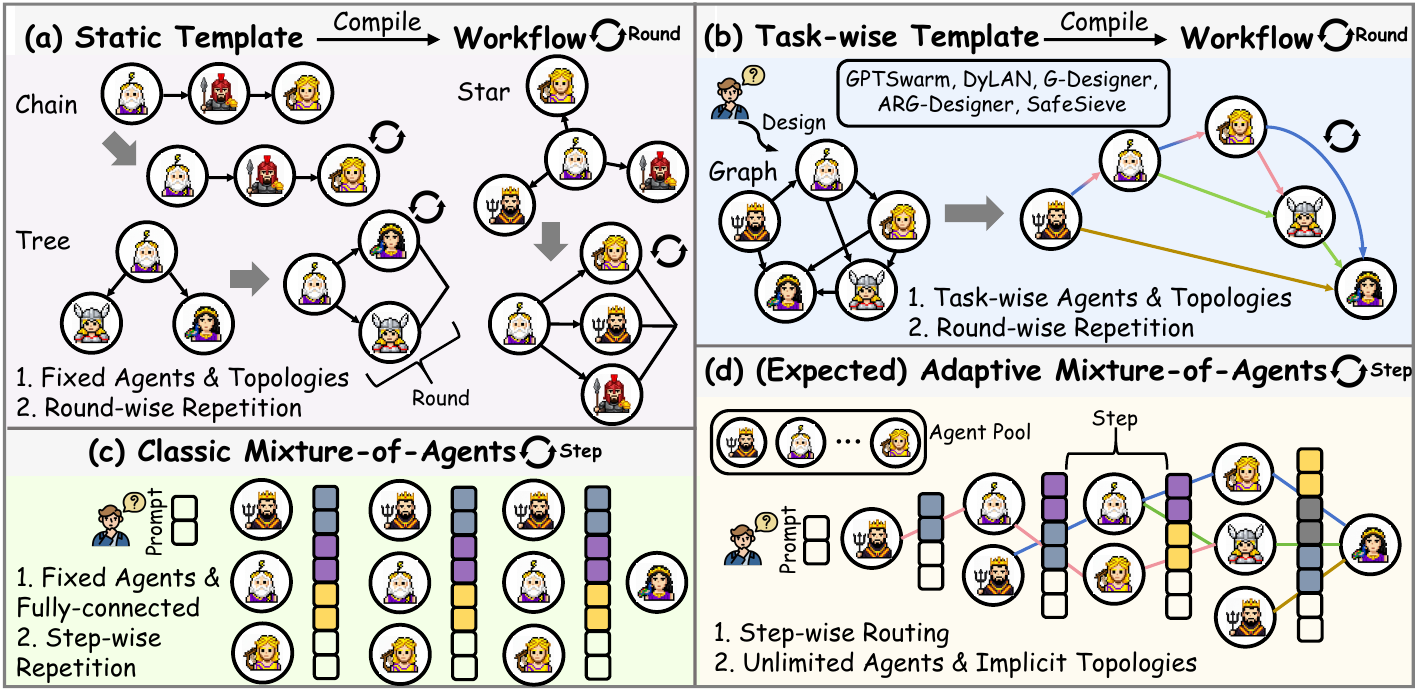}
    \caption{Current Multi-agent Systems (MAS) show inflexibility in different aspects, which have fixed topologies like (a) and (c), or round-wise repetitions like (a) and (b). Though (c) Mixture-of-Agents (MoA) provides the potential of cross-agent interactions in steps, it still follows static structures with fixed numbers of agents and layers. Intuitively, we expect (d) adaptive MoA, which can flexibly determine any number of agents in each step to implicitly simulate arbitrary communication topologies, and adaptively extend with steps, yielding spatio-temporal flexibility. }
\label{fig: intro}
\end{figure*}

In recent years, the surging popularity of Large Language Models (LLM) has stimulated considerable research interests in autonomous agents, which integrates persistent memory and tool utilization with LLMs' strong reasoning capability, to tackle complex tasks like math reasoning~\citep{tora}, code generation~\citep{zheng2023codegeex}, autonomous driving~\citep{jin2023surrealdriver}, and deep research~\citep{zheng2025deepresearcher}. In these scenarios, researchers tend to construct Multi-Agent Systems (MAS)~\citep{hong2023metagpt,autogpt}, aiming to incentivize LLMs' collective intelligence. As shown in Figure~\ref{fig: intro} (a), researchers often pre-define the agent nodes and communication topologies of multi-agent systems. Though they are useful for some specific tasks, the flexibility is constrained when facing versatile scenarios. To overcome this, recent research focuses on ``self-evolving'' MAS~\citep{wang2025evoagentx}, which can adaptively self-adjust to cope with distinct tasks. Here ``self-evolving'' is scoped to agent scheduling, while the candidate agents stay fixed.

Since the communication topologies in MAS are often abstracted as Directed Acyclic Graphs, current self-evolving methods shown in Figure~\ref{fig: intro} (b) mainly focus on adaptively designing query-wise communication topologies through conditional graph generation~\citep{GDesigner,li2025assemble,zhang2025safesieve}. However, this graph-based paradigm still requires to ``pre-compile'' the MAS, where the agent nodes and communication topologies are first determined by a learnable designer, and then compiled into executable topological workflows. In complex scenarios, designing and compiling a MAS based on the query in advance may neglect the potential varying demands during execution, which causes insufficient solutions or excessive token
consumption. Intuitively, we expect an ``elastic'' MAS, which can \textit{extend with varying demands in processing a query, dynamically adjusting the agents and communications in reasoning steps}, instead of predefining workflows with fixed agents, paths, and number of steps.

We draw inspiration from Mixture-of-Agents (MoA)~\citep{wangmixture,DBLP:conf/acl/Xie0SCZWZ25}, a recent multi-agent system. In Figure~\ref{fig: intro} (c), the original MoA also follows the static design by simply concatenating multiple agents' outputs in the current reasoning step, and conveying to all agents in the next reasoning step. Our key insight lies in that, if agents can be spontaneously determined in each reasoning step--see Figure~\ref{fig: intro} (d), the expected Adaptive MoA can directly simulate the workflows of many multi-agent systems, e.g., Chain, Star. For complex topologies such as the one shown in Figure~\ref{fig: intro} (b), the Adaptive MoA in Figure~\ref{fig: intro} (d) can also implicitly simulate it in multiple steps, where reasoning paths of the same color correspond to each other. Compared with graph-based MAS, the expected Adaptive MoA works elastically, which autonomously harness agents and adaptively grow with step granularity, yielding spatio-temporal unboundedness.

To achieve the expected Adaptive MoA, the core challenges lie in the design and optimization of the routing mechanism, which determines what agents should be selected in each reasoning step. Analogous to Mixture-of-Experts (MoE)~\citep{DBLP:conf/iclr/ShazeerMMDLHD17,DBLP:conf/iclr/LepikhinLXCFHKS21}, we design a differentiable routing mechanism for MoA, which considers historical information and the varying contextual information in each reasoning step, then outputs the logits of all candidate agents, and produces the sparse activation. To effectively optimize the differentiable routing mechanism, we introduce a well-recognized endogenous posterior metric, i.e., entropy~\citep{mackay1992practical}, as the supervision signal. Overall, these designs jointly construct the proposed \textbf{\underline{D}}ifferentiable \textbf{\underline{M}}ixture-\textbf{\underline{o}}f-\textbf{\underline{A}}gents (\ourmethod), and we  summarize our core contributions as follows:
\begin{itemize}[leftmargin=2em,itemsep=0em,parsep=0pt,topsep=0pt]
\item[\ding{182}] We introduce \ourmethod, a self-evolving multi-agent system, which adaptively routes agents in each reasoning step based on the contextual information, thus adapting to the varying demands during execution. Essentially, it dynamically adjusts the communication topology along the temporal dimension, which resolves the static compilation dilemma in graph-based multi-agent systems.
\item[\ding{183}] We devise a differentiable context-aware routing mechanism, which routes agents based on the varying contextual information in each reasoning step, where the number of activated agents is also adaptively determined instead of being fixed in advance. We propose to optimize it through the predictive entropy, which can be obtained after executing an agent. This self-supervised paradigm supports \ourmethod with test-time training.
\item[\ding{184}] We conduct comprehensive experiments on 9 benchmarks from multiple real-world tasks, and demonstrate the state-of-the-art performance of \ourmethod. Extensive model analyses on efficiency, robustness, and ensembling capability also demonstrate \ourmethod's elasticity when facing versatile scenarios. The resources are available at \href{https://anonymous.4open.science/r/D-MoA-466C}{https://anonymous.4open.science/r/D-MoA-466C}. 
\end{itemize}

\section{Related works}
\subsection{LLM-powered Multi-Agent Systems}
Due to the success of single-agent systems~\citep{tora,hong2023metagpt,li2025search}, recent interests shift towards multi-agent systems that activate LLMs' collective intelligence. Common approaches predefine communication templates and compile them into fixed executable workflows, such as Chain~\citep{hong2023metagpt}, Tree~\citep{yao2023tree}, Star~\citep{wu2024autogen}, and Graph~\citep{zhuge2024gptswarm}, among which the Directed Acyclic Graph covers the above-mentioned topologies and is widely adopted~\citep{langgraph,wu2024autogen}. This ``design-then-compile'' paradigm requires specialists to decompose tasks in advance, which hinders general multi-agent systems.

\subsection{Self-evolving Multi-Agent Systems}
To support versatile tasks, recent studies turn to automated topology design~\citep{zhuge2024gptswarm,liu2023dynamic,GDesigner}, where graphs serve as the fundamental structure for representing agent interactions~\citep{liu2022temporal,chan2023chateval,hu2024learning}. DyLAN~\citep{liu2023dynamic}, MacNet~\citep{qian2024scaling}, and GPTSwarm~\citep{zhuge2024gptswarm} learn common graph-based topologies for a kind of tasks through iterative optimization, while G-Designer~\citep{GDesigner}, ARG-Designer~\citep{li2025assemble}, and SafeSieve~\citep{zhang2025safesieve} adopt conditional graph generation to design query-wise topologies. Essentially, these studies replace human experts with generative designers, but still suffer the static compilation dilemma, since they predefine the communication rules and neglect the varying demands during execution.

\subsection{Mixture-of-Agents}
Mixture-of-Agents (MoA)~\citep{wangmixture,DBLP:conf/acl/Xie0SCZWZ25} parallelly executes multiple agents in each layer with the same input, then concatenates their outputs with the original prompt and conveys them to the next layer, so that agents from different sources mutually benefit. We observe that it actually constructs fully-connected communication graphs between two reasoning steps. Analogous to Mixture-of-Experts (MoE)~\citep{DBLP:conf/iclr/ShazeerMMDLHD17,DBLP:conf/iclr/LepikhinLXCFHKS21}, such a dense structure maintains full connectivity. We instead equip MoA with a differentiable router, which adaptively routes agents based on the varying contexts.

% Declared one section early so this box reaches the top of the Problem Formalization column.
\begin{table}[!t]
    \centering
    \small
    \begin{tcolorbox}[colback=white,colframe=black,boxrule=0.8pt,arc=2pt,left=4pt,right=4pt,top=2pt,bottom=2pt,before skip=0pt,after skip=0pt]
    \begin{tabular}{@{}p{1\linewidth}@{}}
    \textbf{[Agent Profile]} \\
    You are an expert mathematical analyst. You will be given a math problem, alongside analysis and code from other agents. You must first analyze the problem-solving process step-by-step using algebraic variables. Next, substitute the actual values into your analytical framework to perform calculations and derive the final result. \\
    \\
    \textit{Available Tools:} \\
    $\bullet$ \textcolor{blue}{\texttt{[Calculator]}} evaluates standard mathematical expressions and returns the numerical result.\\
    $\bullet$ \textcolor{ForestGreen}{\texttt{[Python]}} executes Python scripts for complex algorithms or symbolic math, returning the standard output. \\
    $\bullet$ \textcolor{purple}{\texttt{[Search]}} invokes an external search engine to search the corresponding queries on the Internet. \\
    \midrule
    \textbf{[User Query $\mathcal{P}_{usr}$]} \\
    Solve the following mathematical problem: \\
    Alice and Bob take turns rolling a fair 6-sided die, and Alice rolls first. The first person to roll a '6' wins. Calculate the exact probability that Alice wins and the expected number of rolls before the game concludes. \\
    \midrule
    \textbf{[Agent Context \& Synthesis $\mathcal{P}_{last}$]} \\
    You have been provided with preliminary responses from various agents to the user query above. Your task is to synthesize them into a single, logically coherent final resolution that adheres to the highest standards of mathematical accuracy. \\
    \\
    \textit{Responses from agents:} \\
    1. [Agent Response $\mathcal{R}_{i-1,1}$ from $a_{i-1,1}$] \\
    2. [Agent Response $\mathcal{R}_{i-1,2}$ from $a_{i-1,2}$] \\
    ... \\
    $\mathrm{k}_{i-1}$. [Agent Response $\mathcal{R}_{i-1,\mathrm{k}_{i-1}}$ from $a_{i-1,\mathrm{k}_{i-1}}$] \\
    \end{tabular}
    \end{tcolorbox}
    \caption{The prompt to activate agents' capabilities and aggregate the contextual information.}
    \label{tab:template}
\end{table}

\section{Problem Formalization}
This section introduces key concepts and notations in \ourmethod. We model \ourmethod as a step-wise agent-routing system, which routes agents from an initial agent pool $\mathcal{A} = \lbrace a_1, a_2, \cdots, a_\mathrm{N} \rbrace$ with $\mathrm{N}$ agents. Each agent $a_i$ consists of several key elements: $a_i = \lbrace \texttt{LLM}_i, \texttt{Profile}_i, \texttt{Tool}_i \rbrace$, where $\texttt{LLM}_i$ denotes the large language model instance powering $a_i$, $\texttt{Profile}_i$ denotes the system prompt activating $a_i$'s corresponding capability, and $\texttt{Tool}_i$ denotes $a_i$'s external tool set, such as web searcher~\citep{li2025search} and code interpreter~\citep{tora}. 

In reasoning step $i$, the routing mechanism adaptively selects several agents $\lbrace a_{i,1},a_{i,2}, \cdots, a_{i,\mathrm{k}_{i}} \rbrace$ from the pool, where $\mathrm{k}_i < \mathrm{N}$ can be a variable across different steps. Each selected agent $a_{i,j}$ receives prompt $\mathcal{P}_{i,j}$, which consists of two main parts: $\mathcal{P}_{i,j} = (\mathcal{P}_{sys}, \mathcal{P}_{last})$. Among them, the system prompt $\mathcal{P}_{sys} = ( \texttt{Profile}_{i,j}, \mathcal{P}_{usr})$ integrates agent $a_{i,j}$'s profile descriptions and the query $\mathcal{P}_{usr}$ to help recall the problem. Receiving the prompt $\mathcal{P}_{i,j}$, agent $a_{i,j}$ generates response $\mathcal{R}_{i,j}$.  $\mathcal{P}_{last}$ is obtained by concatenating responses $\lbrace \mathcal{R}_{i-1,1}, \mathcal{R}_{i-1,2}, \cdots, \mathcal{R}_{i-1,\mathrm{k}_{i-1}} \rbrace$ from the last reasoning step $i-1$~\citep{wangmixture}. We provide a sample of prompt $\mathcal{P}_{i,j}$ in Table~\ref{tab:template}.

As shown in Table~\ref{tab:template}, $\mathcal{P}_{last}$ considers agents' responses in the last reasoning step as contextual information, and the query $\mathcal{P}_{usr}$ is also provided to activate $a_{i,j}$'s own reasoning. Since the routed agents possess different base LLMs, profiles, or tools, aggregating their responses in this way boosts collaboration. Through step-wise routing, \ourmethod models a self-evolving MAS to construct appropriate communications based on the varying contexts.

\section{Differentiable Mixture-of-Agents}

\begin{figure*}[!htbp]
    \centering
\includegraphics[width=1\linewidth]{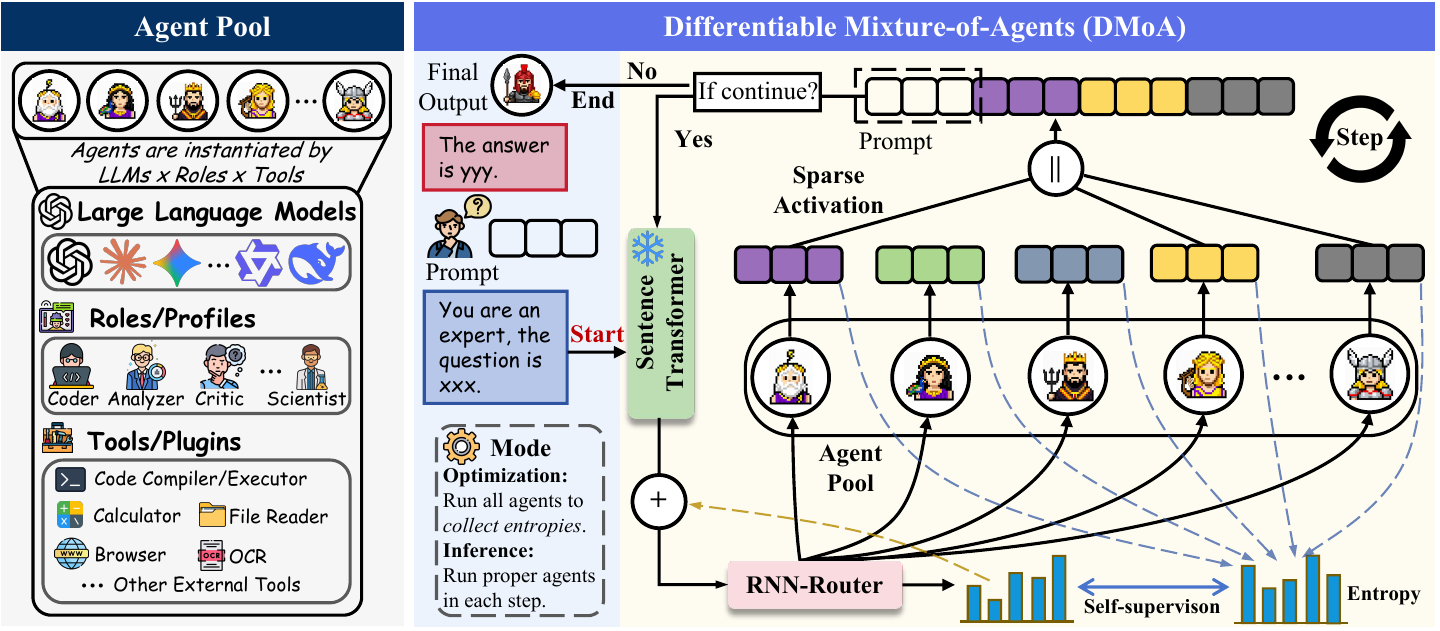}
    \caption{The overview of \ourmethod. An agent pool is initialized to possess diverse expert capabilities. During optimization, \ourmethod runs all agents to collect the predictive entropy, and utilize it as the supervision signal. During inference, only several agents are activated in each reasoning step. }
\label{fig: overview}
\end{figure*}

Figure~\ref{fig: overview} illustrates the structure of \ourmethod. All the routed agents are from a predefined agent pool, where agents possess distinct LLMs, profiles, and tools, to support versatile expert capabilities. Note that agent is a runtime concept, thus predefining agents does not consume physical resources. Compared with original MoA~\citep{wangmixture}, which invokes fixed agents in each step without routing, our proposed \ourmethod is more analogous to the Mixture-of-Experts (MoE)~\citep{DBLP:conf/iclr/ShazeerMMDLHD17} structure in machine learning. Specifically, the core motivation of \ourmethod is that multiple agents specialize in different skill sets, and routing the appropriate agents in a reasoning step can effectively and efficiently address the current step-wise demand. To achieve this, \ourmethod possesses a differentiable routing mechanism like MoE, which can also be optimized through executed results, and supports sparse activation of agents in inference. We further introduce the design details of the differentiable routing mechanism in Section~\ref{sec:routing}, the optimization process of \ourmethod in Section~\ref{sec:optimization}, and the inference process in Section~\ref{sec:inference}.

\subsection{Differentiable Routing Mechanism}
\label{sec:routing}
From a high-level perspective, our proposed \ourmethod extends the MoE concept in reasoning level rather than token level, where the routing action incurs before each reasoning step to determine what agents to execute. Intuitively, the routing mechanism requires to consider: (1) the contextual information, which reflects the progress of query processing and intermediate demands; (2) the historical routing decisions, which helps construct and adjust communications across reasoning steps; (3) how to maintain an optimizable structure with a differentiable router and the explicit objectives.  

Specifically, to fulfill the above-mentioned requirements, we design the differentiable routing mechanism using a lightweight Sentence Transformer~\citep{reimers1908sentence}, denoted as $\texttt{ST}(\cdot)$, and a recurrent neural network~\citep{cho2014learning}, denoted as $\texttt{RNN}(\cdot)$. The Sentence Transformer is a bidirectional Transformer language model, used to compress contextual information into a fixed-size semantic vector. The recurrent neural network is used to consider the historical routing rationales to construct cross-layer communications, and support autoregressive extension to future reasoning steps during inference. Given query $\mathcal{P}_{usr}$, we initialize the hidden state as $\mathrm{h}_0=\mathbf{0}$ and the semantic center as $\mathrm{X}_1 = \texttt{ST}(\mathcal{P}_{usr})$, and formulate the routing in the $i$-th reasoning step as:
\begin{gather}
    \mathrm{h}_{i}, \mathrm{o}_i = \texttt{RNN}(\mathrm{X}_i, \mathrm{h}_{i-1}),\ \ \mathrm{z}_i = \texttt{Linear}(\mathrm{o}_i),\\
    \mathrm{k}_i = \texttt{Min}\left(\mathrm{K},\ \texttt{Count}(\tilde{\mathrm{z}}_i \geq 1/\mathrm{N})\right), \label{eq:topk}\\
    \mathcal{A}_i = \texttt{TopK}(\mathrm{z}_i, \mathcal{A}, \mathrm{k}_i),\ \ \mathcal{R}_{i,j} = a_{i,j}(\mathcal{P}_{i,j}), \label{eq:select}
\end{gather}
where $\mathrm{X}_i\in\mathbb{R}^D$ is the semantic center of the contexts, $\mathrm{h}_i\in\mathbb{R}^D$ is the compressed hidden state serving as the historical information for subsequent routing, and the $\texttt{RNN}$ output $\mathrm{o}_i\in\mathbb{R}^D$ is further converted to logits $\mathrm{z}_i \in \mathbb{R}^{\mathrm{N}}$, where $\mathrm{N}$ is the size of the agent pool $\mathcal{A}$. We write $\tilde{\mathrm{z}}_i = \texttt{Softmax}(\mathrm{z}_i/\tau)$ for the normalized routing weights with temperature $\tau$, and $\mathcal{A}_i = \lbrace a_{i,1}, \cdots, a_{i,\mathrm{k}_i} \rbrace$ for the agents routed in step $i$, which $\texttt{TopK}$ picks from $\mathcal{A}$ by the largest $\mathrm{k}_i$ logits in $\mathrm{z}_i$. The number $\mathrm{k}_i$ is adaptively determined through Equation~(\ref{eq:topk}): an agent is kept only if its normalized weight is (1) greater than the average value $1/\mathrm{N}$ and (2) among the top $\mathrm{K}$ ones. Therefore, \ourmethod can adaptively routes no more than $\mathrm{K}$ agents in each reasoning step, and adjusting the predefined $\mathrm{K}$ and the temperature $\tau$ helps control the routing sensitivity. Each routed agent is executed with prompt $\mathcal{P}_{i,j}$, which follows the prompt template in Table~\ref{tab:template}. On the other hand, to enable the routing mechanism to perceive changes in contextual information and their magnitudes, we weight-sum the semantic features of the routed agents' responses into the semantic center of the next reasoning step:
\begin{gather}
    \hat{\mathcal{R}}_{i,j} = \texttt{ST}(\mathcal{R}_{i,j}), \label{eq:8}\\
    \boldsymbol{\alpha}_{i} = \texttt{Softmax}(\texttt{TopK}(\mathrm{z}_{i},\mathrm{k}_{i})),\\
    \mathrm{X}_{i+1} = \textstyle\sum_{j=1}^{\mathrm{k}_{i}} \alpha_{i}^{(j)} \hat{\mathcal{R}}_{i,j}. \label{eq:9}
\end{gather}
Based on the above-mentioned designs, the gradient flow can be propagated across the multiple reasoning steps through the differentiable RNN-Router, and weights $\boldsymbol{\alpha}_{i}$, thus constructing the Differentiable Mixture-of-Agents.

% Declared one page early so this full-width table reaches the top of the Main Results page.
\begin{table*}[t]
    \centering
    \renewcommand\tabcolsep{5.3pt}
    \renewcommand\arraystretch{1.1}
    \resizebox{\linewidth}{!}{
    \begin{tabular}{lcccccccccc}
    \toprule
    \rowcb
        \textbf{Methods}& \textbf{MMLU} & \textbf{GSM8K} & \textbf{MultiArith} & \textbf{SVAMP} & \textbf{AQuA} & \textbf{HumanEval} & \textbf{DS-1000} & \textbf{HotpotQA} & \textbf{DDXPlus} & \textbf{Avg.} \\ \midrule
        Vanilla & 80.47  & 87.15  & 93.40  & 87.25  & 69.27  & 73.28  & 38.40  & 82.35  & 56.40  & 74.22   \\ \hline
        \multicolumn{11}{l}{\textit{\textbf{\small Single-Agent Systems}}}
        \\
        \rowcg
        CoT  & 81.85\red{1.38}  & 87.35\red{0.20}  & 93.60\red{0.20}  & 87.90\red{0.65}  & 72.40\red{3.13}  & 74.05\red{0.77}  & 42.35\red{3.95}  & 81.04\blue{1.31}  & 64.22\red{7.82}  & 76.08   \\ 
        SC & 82.95\red{2.48}  & 87.66\red{0.51}  & 94.40\red{1.00}  & 87.40\red{0.15}  & 71.85\red{2.58}  & 75.82\red{2.54} & 37.66\blue{0.74}  & 82.24\blue{0.11}  & 66.43\red{10.03}  & 76.27   \\ \hline
        \multicolumn{11}{l}{\textit{\textbf{\small Multi-Agent Systems}}}
        \\
        \rowcg
        Chain  & 82.40\red{1.93}  & 87.23\red{0.08}  & 92.88\blue{0.52}  & 87.16\blue{0.09}  & 69.85\red{0.58}  & 73.40\red{0.12}  & 38.25\blue{0.15}  & 80.77\blue{1.58}  & 65.28\red{8.88}  & 75.25   \\ 
        Tree  & 81.55\red{1.08}  & 86.35\blue{0.80} & 92.95\blue{0.45}  & 88.25\red{1.00}  & 71.44\red{2.17}  & 75.33\red{2.05} & 43.50\red{5.10}  & 81.50\blue{0.85} & 69.25\red{12.85}  & 76.68   \\ 
        \rowcg
        Star  & 79.64\blue{0.83}  & 85.25\blue{1.90}  & 93.25\blue{0.15}  & 87.60\red{0.35}  & 68.50\blue{0.77}  & 76.28\red{3.00}  & 41.82\red{3.42}  & 83.52\red{1.17}  & 68.25\red{11.85}  & 76.01   \\ 
        Complete Graph  & 82.75\red{2.28}  & 88.05\red{0.90}  & 94.73\red{1.33}  & 88.20\red{0.95}  & 70.44\red{1.17}  & 84.25\red{10.97}  & 45.05\red{6.65}  & 83.35\red{1.00}  & 70.35\red{13.95}  & 78.57   \\ 
        \rowcg
        Random Graph & 83.20\red{2.73}  & 88.25\red{1.10}  & 94.80\red{1.40}  & 87.10\blue{0.15}  & 70.16\red{0.89}  & 84.14\red{10.86}  & 47.48\red{9.08}  & 84.28\red{1.93}  & 70.44\red{14.04}  & 78.87   \\ 
        AutoGen & 83.25\red{2.78}  & 89.40\red{2.25}  & 95.05\red{1.65}  & 88.15\red{0.90}  & 71.50\red{2.23}  & 83.50\red{10.22}  & 48.55\red{10.15}  & 83.25\red{0.90}  & 71.83\red{15.43}  & 79.39   \\ 
        \rowcg
        LLM-Debate & 83.28\red{2.81}  & 89.95\red{2.80}  & 95.45\red{2.05}  & 88.30\red{1.05} & 73.52\red{4.25}  & 85.24\red{11.96}  & 49.26\red{10.86}  & 84.55\red{2.20}  & 71.33\red{14.93}  & 80.10   \\ 
        MoA & 82.65\red{2.18}  & 88.52\red{1.37} & 94.50\red{1.10} & 88.24\red{0.99} & 67.35\blue{1.92} & 82.48\red{9.20} & 39.75\red{1.35} & 84.02\red{1.67} & 71.85\red{15.45} & 77.71\\
        \hline
        \multicolumn{11}{l}{\textit{\textbf{\small Self-Evolving Multi-Agent Systems}}}
        \\
        \rowcg
        GPTSwarm & 83.80\red{3.33}  & 90.20\red{3.05}  & 96.28\red{2.88}  & 87.28\red{0.03}  & 75.44\red{6.17}  & 87.14\red{13.86}  & 52.35\red{13.95}  & 85.40\red{3.05}  & 75.57\red{19.17}  & 81.50   \\ 
        G-Designer  & 84.63\red{4.16}  & 93.35\red{6.20}  & 97.60\red{4.20}  & 89.52\red{2.27}  & 76.30\red{7.03}  & 89.40\red{16.12}  & 54.82\red{16.42}  & 87.66\red{5.31}  & 76.42\red{20.02}  & 83.30   \\ 
        \rowcg
        ARG-Designer & \second{85.04}\red{4.57}  & \second{94.40}\red{7.25}  & 97.35\red{3.95}  & 88.95\red{1.70}  & 79.48\red{10.21}  & 88.25\red{14.97}  & \second{55.60}\red{17.20}  & 86.45\red{4.10}  & 76.03\red{19.63}  & 83.51   \\ 
        SafeSieve & 84.65\red{4.18}  & 93.20\red{6.05}  & \second{97.80}
        \red{4.40} & \second{90.10}\red{2.85}  & 80.40\red{11.13}  & 90.15\red{16.87}  & 53.73\red{15.33}  & 87.22\red{4.87}  & \second{77.94}\red{21.54}  & \second{83.91}   \\
        \rowcg
        AFlow & 83.22\red{2.75}  & 89.50\red{2.35}  & 94.28\red{0.88}  & 86.48\blue{0.77} & 78.20\red{8.93}  & 89.25\red{15.97}  & 46.82\red{8.42}  & \second{90.04}\red{7.69}  & 74.25\red{17.85}  & 81.34   \\ 
        SpecReason & 84.20\red{3.73}  & 91.40\red{4.25}  & 95.40\red{2.00}  & 87.62\red{0.37}  & 81.36\red{12.09}  & \second{91.22}\red{17.94}  & 50.08\red{11.68}  & 88.29\red{5.94}  & 76.58\red{20.18}  & 82.91   \\ 
        \rowcg
        STEER & 84.40\red{3.93}  & 93.50\red{6.35}  & 95.82\red{2.42}  & 87.40\red{0.15}  & \second{82.66}\red{13.39}  & 89.02\red{15.74}  & 51.34\red{12.94}  & 89.35\red{7.00}  & 75.20\red{18.80}  & 83.19   \\ \hline
        \rowcr
        \ourmethod (Ours)  & \first{91.35}\red{10.88} & \first{98.87}\red{11.72} & \first{99.15}\red{5.75} & \first{94.76}\red{7.51} & \first{86.60}\red{17.33} & \first{95.62}\red{22.34} & \first{64.34}\red{25.94} & \first{90.38}\red{8.03} & \first{83.37}\red{26.97} & \first{89.38} \\ 
        \bottomrule
    \end{tabular}}
    \caption{Accuracy comparison with three types of baselines, including Single-Agent Systems, Multi-Agent Systems, Self-Evolving Multi-Agent Systems. All baselines and \ourmethod are driven by \llmname{gpt-oss-120b} (Vanilla). \first{Red}: the best, \second{Blue}: the runner-up.}
    \label{tab: main results}
\end{table*}

\subsection{The Optimization of \ourmethod}
\label{sec:optimization}
Recent self-evolving multi-agent systems~\citep{GDesigner,li2025assemble,pesce2023learning} adopt reinforcement learning (RL) to optimize their topology designers, since direct gradient propagation is intractable in such complex systems with multiple agents and multi-step interactive reasoning. However, the reward signals are sparse, often outcome rewards obtained through rolling out multi-agent systems for dozens of episodes, which hinders the convergence. Under this paradigm, these self-evolving multi-agent systems also do not support zero-shot generalization to unseen scenarios, which requires few-shot RL finetuning to adapt the topology designers. 

In \ourmethod, the complex interactive reasoning process is decomposed into step-wise granularity, and the self-evolving paradigm relies on the step-wise agent routing, of which the objective is to choose the most appropriate agents to handle the step-wise emergent demands. Since recent studies~\citep{arpo,aepo} point out that entropy~\citep{mackay1992practical} serves as a key indicator to evaluate the confidence of agents, we adopt it to optimize the routing mechanism in a self-supervised proximal optimization paradigm. Given the previously generated tokens $x_{<j}$, an agent generates the next token $x_j$ according to the conditional distribution $p_j(\cdot) \triangleq \mathrm{P}_\Theta(\cdot \mid x_{<j})$. We further define the predictive entropy (PE) as the average conditional entropy over all $M$ tokens:
\begin{equation}
\mathrm{PE} = -\frac{1}{M} \sum_{j=1}^{M} \sum_{x_j} p_j(x_j) \log p_j(x_j).
\end{equation}
During optimization, we simply run all $\mathrm{N}$ agents in the $i$-th reasoning step to collect the prediction entropy of them: $\mathcal{E}_i \in \mathbb{R}^\mathrm{N}$. We then normalize the negative entropy through softmax to reflect the confidence of agents: $\mathcal{C}_i = \texttt{Softmax}(-\mathcal{E}_i) \in \mathbb{R}^{\mathrm{N}}$. We then adopt the pair-wise ranking loss~\citep{burges2005learning} to align the ordering of routing logits $\mathcal{Z}_i = \texttt{Softmax}(\mathrm{z}_i)$ with $\mathcal{C}_i$:
\begin{equation}
\mathcal{L}_{\text{rank}}^{(i)} = \sum_{(a,b) \in \mathcal{S}_i} \log\big(1 + e^{-\delta_i^{(a,b)}}\big),
\end{equation}
where $\mathcal{S}_i = \lbrace (a,b) : \mathcal{C}_i^{(a)} > \mathcal{C}_i^{(b)} \rbrace$ collects the agent pairs ordered by confidence, and $\delta_i^{(a,b)} = \mathcal{Z}_i^{(a)} - \mathcal{Z}_i^{(b)}$ is the corresponding logit margin. The total loss $\mathcal{L}_{\text{rank}}$ across all $L$ reasoning steps is calculated as:
\begin{gather}
\mathcal{L}_{\text{rank}} 
= \frac{1}{L} \sum_{i=1}^{L} \mathcal{L}_{\text{rank}}^{(i)},
\end{gather}
optimized by this objective, the routing mechanism tends to select agents with strong confidence, which may excel at handling current task demands. This self-supervised paradigm has sufficient supervision signals and originally supports test-time training to adapt to zero-shot scenarios.

\subsection{The Inference of \ourmethod}
\label{sec:inference}
\ourmethod supports the flexible extension of reasoning steps and can adaptively route agents at each step, resulting in a communication topology without spatial-temporal boundness. During inference, \ourmethod supports sparse activation to adaptively route $\mathrm{k}_i$ agents in the $i$-th reasoning step. To control the token consumption, we can predefine a maximum number of reasoning steps, and employ a summarizer agent to organize the final answer upon reaching the limit. As an alternative, the summarizer agent can autonomously determine at which step to terminate the reasoning process.

\section{Experiments}
\label{exp}

% main experiments

% efficiency

% ensembling

% test time training DMoA (few-shot) DMoA (TTT) DMoA (few-shot + TTT)

% robustness

% ablation studies on routing mechanism
% LLM-based Selector
% Linear-based Router
% w/o Semantic Aggregation
% w/o Adaptive k_i

% visualized analyses 
% different routing mechanism
% different objective loss

% scalability  (N, k)

% case study

\subsection{Experimental Settings}
\noindent\textbf{Benchmarks.}
We assess \ourmethod on 9 representative benchmarks spanning diverse reasoning settings. Specifically, we use MMLU~\citep{hendrycksmeasuring} to evaluate broad-domain reasoning; GSM8K~\citep{cobbe2021training}, MultiArith~\citep{roy2015solving}, SVAMP~\citep{patel2021nlp}, and AQuA~\citep{ling2017program} for mathematical reasoning; HumanEval~\citep{chen2021evaluating} and DS-1000~\citep{lai2023ds} for coding ability; HotpotQA~\citep{yang2018hotpotqa} for web-based question answering; and DDXPlus~\citep{fansi2022ddxplus} for medical diagnosis reasoning.

\noindent\textbf{Baselines.}
We compare \ourmethod against a broad spectrum of multi-agent systems, including static methods such as COT~\citep{wei2022chain}, Self-Consistency~\citep{wangself}, Chain, Tree, Star, Complete Graph, Random Graph, AutoGen~\citep{wu2024autogen}, MoA~\citep{wangmixture}, and LLM-Debate~\citep{du2023improving}; spatially adaptive methods including GPTSwarm~\citep{zhuge2024gptswarm}, G-Designer~\citep{GDesigner}, ARG-Designer~\citep{li2025assemble}, and SafeSieve~\citep{zhang2025safesieve}; and temporally adaptive methods such as AFlow~\citep{zhangaflow}, SpecReason~\citep{damani2024learning}, and STEER~\citep{lee2025confidence}.

\noindent\textbf{Implementation Details.}
We instantiate \ourmethod and all baselines using \llmname{gpt-oss-120b} as the backbone model. To aggregate intermediate dialogue trajectories and produce the final response, we introduce a summarizer agent. For graph-based MAS, we adopt the proper dialogue iterations in their original papers. Since \ourmethod supports step-wise harnessing, we leverage the summarizer agent to adaptively determine the termination.  For Sentence Transformer, we employ \llmname{all-MiniLM-L6-v2} with embedding size $D=384$. For RNN-Router, we adopt the classic GRU. 

\subsection{Main Results}

We compare \ourmethod with previously advanced multi-agent systems in Table~\ref{tab: main results}. For all datasets, we adopt a commonly-used few-shot paradigm~\citep{GDesigner,li2025assemble}, which uses 40--80 queries from the training sets to train optimizable baselines such as G-Designer, ARG-Designer, and our proposed \ourmethod. Results demonstrate that \ourmethod achieves consistent state-of-the-art performance across 9 benchmarks, with average accuracy improvements (against Vanilla) from \textit{5.75} to \textit{26.97}. Compared with Self-Evolving Multi-Agent Systems, \ourmethod possesses more stable performance on diverse tasks, from simple ones on GSM8k (11.72 $\uparrow$ ) and MMLU (10.88 $\uparrow$), to hard ones on DS-1000 (25.94 $\uparrow$) and DDXPlus (26.97 $\uparrow$). The larger improvements on harder tasks demonstrate the effectiveness of \ourmethod's routing mechanism in dynamically tuning the communications and scheduling the resources.

\subsection{Method Analyses}
\begin{table}[t]
    \centering

    \resizebox{\linewidth}{!}{
    \begin{tabular}{lccccc}
        \toprule
        \rowcb
        Methods & MMLU & GSM8K & HumanEval & DS-1000 & HotpotQA \\
        \midrule
        Vanilla & 80.47 & 87.15 & 73.28 & 38.40 & 82.35 \\
        Random Graph & 83.20 & 88.25 & 84.14 & 47.48 & 84.28 \\
        MoA & 82.65 & 88.52 & 82.48 & 39.75 & 84.02 \\
        \midrule
        GPTSwarm & 83.80 & 90.20 & 87.14 & 52.35 & 85.40 \\
        G-Designer & 84.63 & 93.35 & 89.40 & 54.82 & 87.66 \\
        ARG-Designer & 85.04 & 94.40 & 88.25 & 55.60 & 86.45 \\
        SafeSieve & 84.65 & 93.20 & 90.15 & 53.73 & 87.22 \\
        \midrule
        \rowcg
        DMoA (Few-shot) & 91.35 & 98.87 & 95.62 & 64.34 & 90.38 \\
        \rowcg
        DMoA (TTT) & 91.80 & 98.65 & 96.04 & 65.44 & 89.50 \\
        \rowcg
        DMoA (Few-shot + TTT) & 93.50 & 99.30 & 97.52 & 65.55 & 91.40 \\
        \bottomrule
    \end{tabular}}
    \caption{We evaluate \ourmethod's ``Test Time Training'' capability on MMLU, GSM8K, HumanEval, DS-1000, and HotpotQA, where we collect the predictive entropy on 10--30 queries during inference, and optimize the \ourmethod to adapt to the tasks.  }
    \label{tab:ttt}
\end{table}

\begin{table}[t]
    \centering
    \resizebox{\linewidth}{!}{
    \begin{tabular}{lccccc}
        \toprule
        \rowcb
        Methods & MMLU & GSM8K & HumanEval & DS-1000 & HotpotQA \\
        \midrule
        GPT 5.2 & 83.72 & 88.56 & 79.32 & 42.39 & 84.35 \\
        DeepSeek V3.2 & 85.91 & 92.63 & 84.26 & 36.24 & 79.33 \\
        Qwen3 Max & 83.26 & 88.25 & 82.17 & 38.49 & 81.25 \\
        Gemini 3 Pro & 86.22 & 86.82 & 85.48 & 44.50 & 84.90 \\
        \midrule
        Qwen3 32B Dense* & 57.68 & 74.25 & 62.32 & 24.75 & 56.32 \\
        Llama3 70B Dense* & 41.28 & 65.32 & 66.38 & 28.75 & 49.65 \\
        DeepSeek V3 67B Dense* & 49.56 & 70.20 & 58.66 & 32.50 & 44.76 \\
        Mistral 8x22B v0.1* & 62.45 & 72.30 & 60.35 & 22.05 & 46.67 \\
        \midrule
        MoA* & 72.54 & 80.47 & 77.32 & 36.52 & 69.49 \\
        \rowcg
        DMoA (Few-shot)* & 83.65 & 85.48 & 83.27 & 43.55 & 81.87 \\
        \rowcg
        DMoA (TTT)* & 83.74 & 86.27 & 84.80 & 43.74 & 82.25 \\
        \rowcg
        DMoA (Few-shot + TTT)* & 85.52 & 88.74 & 86.72 & 48.97 & 86.50 \\
        \bottomrule
    \end{tabular}}
    \caption{We evaluate \ourmethod's ensembling capability on MMLU, GSM8K, HumanEval, DS-1000, and HotpotQA, where we utilize the open-source LLMs (rows 5--8) in \ourmethod, and compare its performance with closed-source strong LLMs (rows 1--4). * denotes open-source LLMs or MAS driven by open-source LLMs.}
    \label{tab:ensemble}
\end{table}

\noindent\textbf{Test Time Training.} \ourmethod is optimized through self-supervision signals from the step-wise predictive entropy, which is dense and easily obtainable, thus \ourmethod originally supports test time training. Specifically, on the first 10--30 queries in the test set of each benchmark, we run all $\mathrm{N}$ agents in \ourmethod to collect the predictive entropy as the supervision signal, then optimize the routing mechanism to adapt to the current task, and continue sparse activation for the remaining queries. 

We provide three variants of \ourmethod in Table~\ref{tab:ttt}, where \ourmethod (Few-shot) follows the few-shot paradigm in Table~\ref{tab: main results}, \ourmethod (TTT) follows the above-mentioned test time training paradigm, and \ourmethod (Few-shot + TTT) integrates both of them. Results show that \ourmethod (Few-shot + TTT) achieves best performance as expected, and \ourmethod (TTT) achieves competitive performance against \ourmethod (Few-shot), which demonstrates \ourmethod's adaptive capabilities when facing tasks of different categories. This provides a solution for lifelong self-evolving MAS. 

\noindent\textbf{Ensembling capability.} As introduced in MoA~\citep{wangmixture}, heterogeneous LLMs can obtain complementary information from their respective outputs, thus possessing ensembling capabilities to achieve competitive performance against closed-source LLMs with hundreds of billions parameters. 

In \ourmethod, this capability is preserved and amplified. As shown in Table~\ref{tab:ensemble}, we instantiate \ourmethod with the \textit{open-source} LLMs in rows 5--8, and allow all baselines to utilize the tools in \ourmethod for a fair comparison. It is observed that \ourmethod achieves even better performance than \textit{closed-source} LLMs like GPT 5.2 and Gemini 3 Pro on most settings. This demonstrate \ourmethod can effectively incentivize LLMs' swarm intelligence.

\begin{figure}[!ht]
    \centering
    \includegraphics[width=\columnwidth]{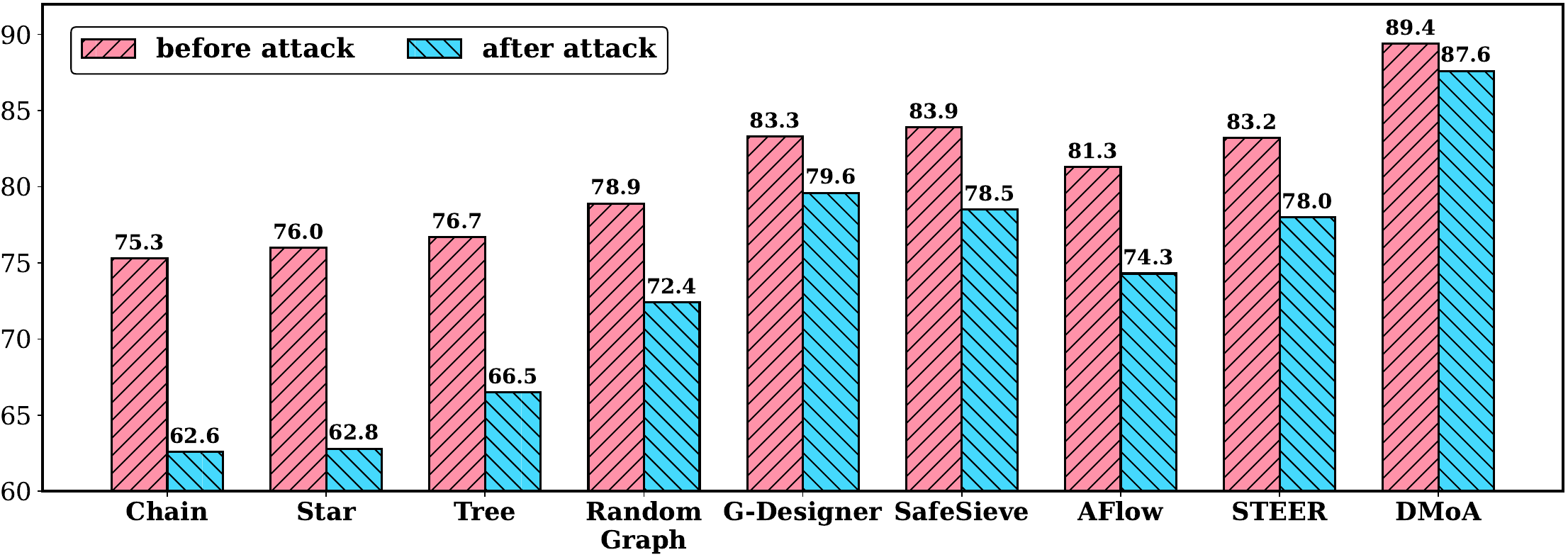}
    \caption{Analyses of robustness. We compare the accuracy (\%) of multiple multi-agent systems before and after prompt attacks on all benchmarks, and report the average accuracies.}
    \label{fig:robustness}
\end{figure}

\noindent\textbf{Robustness analyses.} We further evaluate the robustness of \ourmethod under external perturbations. Following the prompt injection setting introduced in G-Designer~\citep{GDesigner}, we compare \ourmethod with representative baselines in Figure~\ref{fig:robustness}. The results indicate that static multi-agent systems are notably vulnerable to partial system attacks, exhibiting accuracy drops of \textit{6.5\%--12.7\%}. In contrast, self-evolving methods such as G-Designer and SafeSieve are more resilient due to their task-aware topology adaptation, with performance degradation remaining below 5\%. Benefiting from the same adaptive topology mechanism, \ourmethod demonstrates even stronger robustness under adversarial perturbations, maintaining a slight decrease before and after the attacks.

\begin{figure}[t]
    \centering
    \includegraphics[width=\columnwidth]{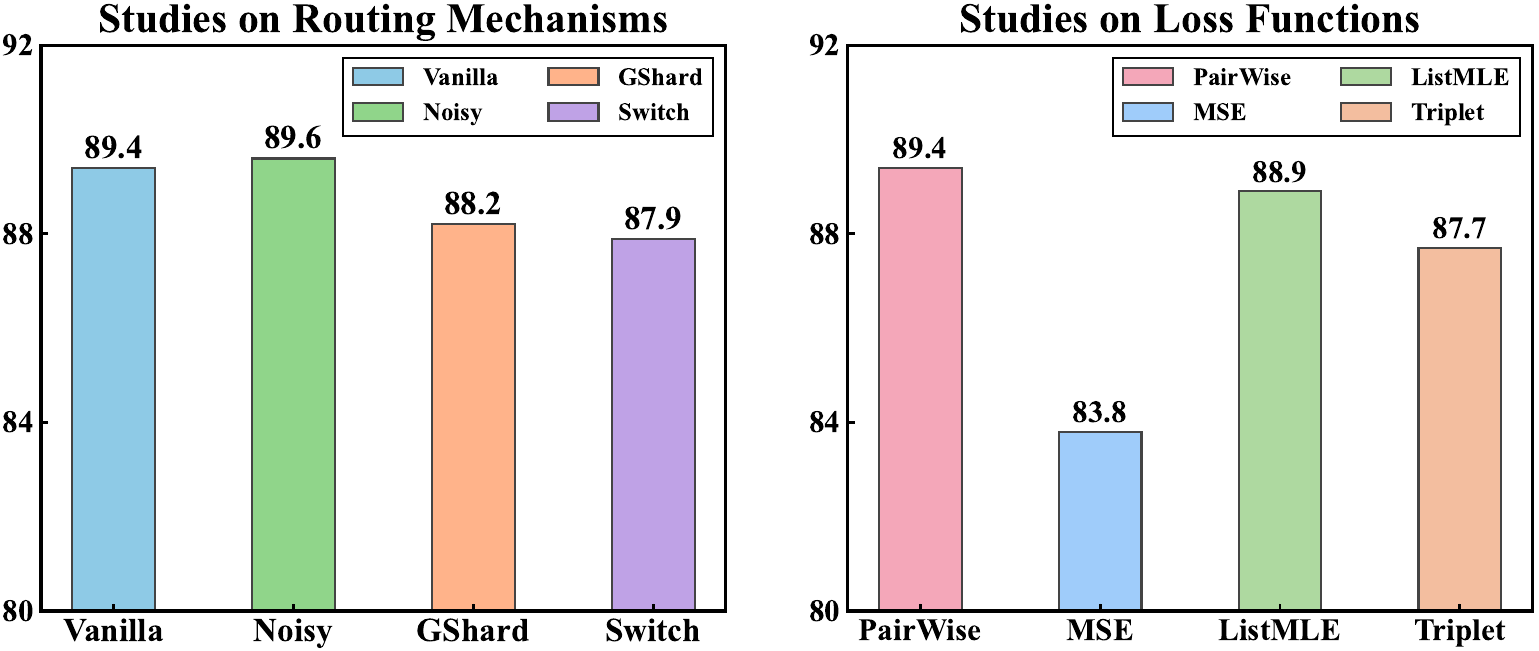}
    \caption{Comparisons among different routing mechanisms and loss functions on all benchmarks.}
    \label{fig:routing_mechanism}
\end{figure}
\begin{table}[!ht]
    \centering

    %==================== Table ====================
    \resizebox{\linewidth}{!}{
    \begin{tabular}{lccccc}
        \toprule
        \rowcb
        Variants & MMLU & GSM8K & HumanEval & DS-1000 & HotpotQA \\
        \midrule
        w LLM Selector & 85.56 & 92.84 & 87.21 & 54.33 & 84.72 \\
        w Linear Router & 89.76 & 96.84 & 92.11 & 60.02 & 88.71 \\
        w/o Aggregation & 89.58 & 96.67 & 91.94 & 59.86 & 88.83 \\
        w/o Adaptive $\mathrm{k}_i$ & 90.47 & 97.58 & 93.26 & 61.72 & 89.41 \\
        \rowcg
        DMoA & \textbf{91.35} & \textbf{98.87} & \textbf{95.62} &\textbf{ 64.34} & \textbf{90.38} \\
        \bottomrule
    \end{tabular}}
    \caption{Ablation studies on key components.}
    \label{tab:ablation}
\end{table}

\noindent\textbf{Ablation studies.} We conduct ablation studies in Table~\ref{tab:ablation}: (1) w LLM Selector, replacing the whole routing mechanism with GPT 5.2; (2) w Linear Router, replacing the RNN-Router with classic linear router~\citep{DBLP:conf/iclr/ShazeerMMDLHD17}; (3) w/o Aggregation, removing the aggregation process in Equation (\ref{eq:8}--\ref{eq:9}), and directly extracting the semantic vectors of the contextual information; (4) w/o Adaptive $\mathrm{k}_i$, fixing the routing number to $\mathrm{K}$. Results demonstrate the consistent effectiveness of all key components.

We further analyze alternative routing mechanisms from Mixture-of-Experts (MoE), i.e., Vanilla (Ours), Noisy Gating~\citep{DBLP:conf/iclr/ShazeerMMDLHD17}, GShard~\citep{DBLP:conf/iclr/LepikhinLXCFHKS21}, and Switch~\citep{zhou2022mixture}. Results in Figure~\ref{fig:routing_mechanism} left show that the Vanilla or noisy gating achieve better performance. Since \ourmethod has sufficient supervision signals and does not suffer expert collapse, GShard and Switch lose their advantages. Considering loss functions, results in Figure~\ref{fig:routing_mechanism} right also demonstrate the advantages of pair-wise ranking loss over MSE, ListMLE~\citep{xia2008listwise}, and Triplet~\citep{schroff2015facenet}. 

\begin{figure}[t]
    \centering
    \includegraphics[width=\columnwidth]{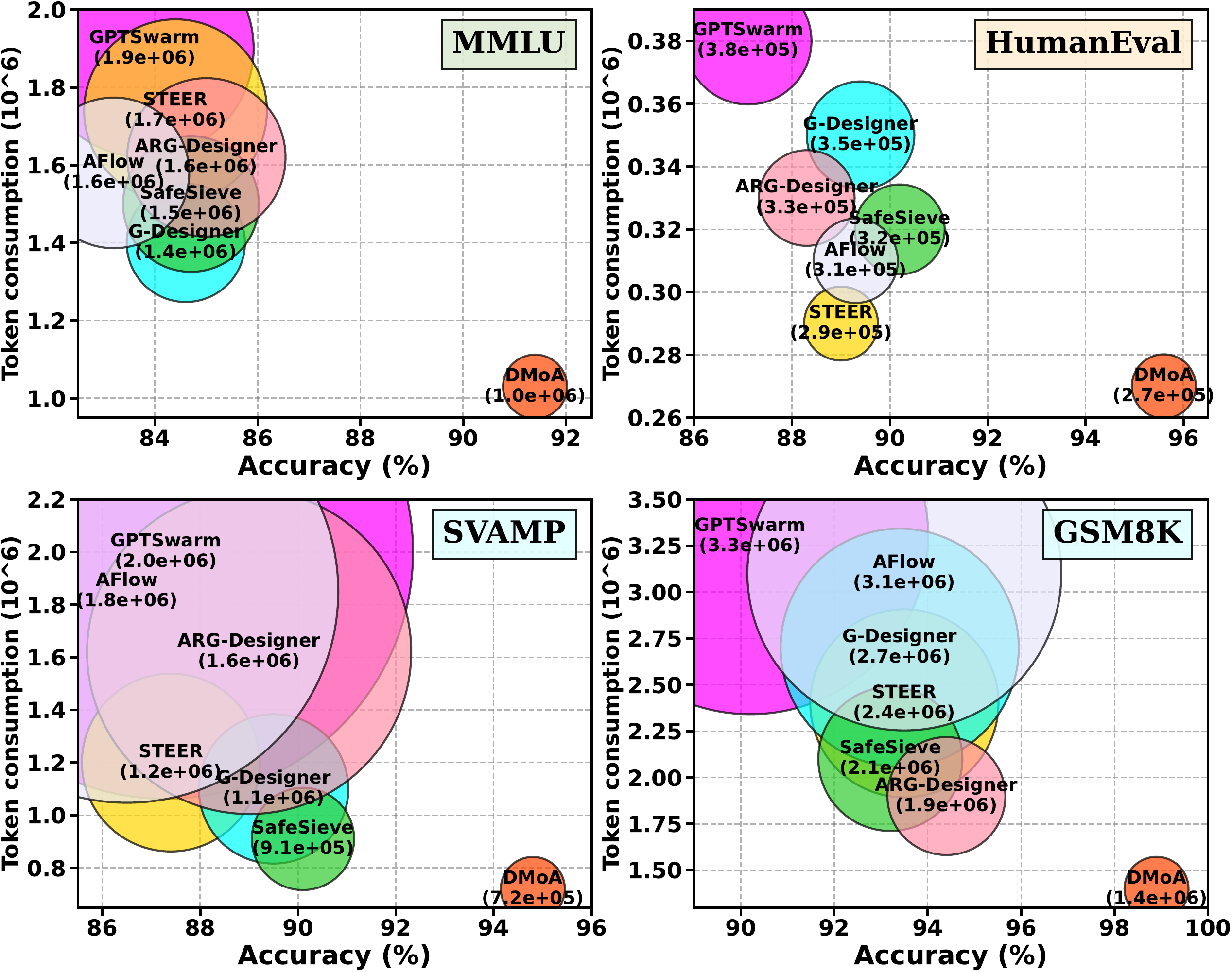}
    \caption{Visualization of accuracy and consumption of multi-agent systems across MMLU, HumanEval, GSM8K, and SVAMP. The diameters of circles represent the scales of token consumption.}
    \label{fig:efficiency}
\end{figure}

\noindent\textbf{Efficiency Analyses.}  We examine the efficiency of \ourmethod in Figure~\ref{fig:efficiency}. For a fair comparison, we ensure the same agent budget for all multi-agent systems: for graph-based MAS, we adopt a fixed number of agents for the same benchmark; for \ourmethod, we keep the maximum agent routing number $\mathrm{K}$ equal to that number. Results show that \ourmethod consistently achieves substantially stronger performance than all baselines while consuming fewer tokens. It is suggested that \ourmethod can elastically coordinate computational resources and activate specialized agents only when they are truly needed. In comparison with recent strong baselines such as ARG-Designer and SafeSieve, \ourmethod attains at least $5.47$ higher accuracy on MMLU and HumanEval while requiring far less token usage. 

\noindent\textbf{Scalability Analyses.} The scalability of \ourmethod is mainly affected by the agent pool size $\mathrm{N}$ and the predefined maximum routing number $\mathrm{K}$, where a larger $\mathrm{N}$ provides more diverse agents, and a larger $\mathrm{K}$ introduces more cross-step communications. Enumerating multiple $(\mathrm{N}, \mathrm{K})$ configurations in Figure~\ref{fig:scalability} and Table~\ref{tab:scalability_nk_appendix} of Appendix~\ref{app:scalability}, we observe consistent accuracy improvements on all benchmarks, which demonstrates the scalability of \ourmethod.

\section{Conclusion}
We introduce \ourmethod, a self-evolving multi-agent system that elastically extends the communication topology based on the varying task demands, yielding spatio-temporal flexibility. Instead of pre-compiling a workflow, \ourmethod routes agents at the step granularity, and optimizes the router with the predictive entropy of the executed agents. Experiments on nine benchmarks show that \ourmethod attains state-of-the-art accuracy with fewer tokens and remains robust under adversarial agents, jointly enabling lifelong swarm intelligence.

\section*{Limitations}
Although \ourmethod achieves strong performance across nine benchmarks, several limitations remain. First, \ourmethod relies on a pool of pre-specified candidate agents; its effectiveness is therefore bounded by the capability and diversity of the underlying LLMs available in the pool, and extending the pool incurs additional inference cost. Second, the differentiable routing and the predictive-entropy objective introduce extra forward passes at test time, which, while efficient relative to exhaustive topology search, still adds overhead compared with a single-agent baseline. Third, our evaluation focuses on reasoning-style benchmarks; generalization to open-ended generation, long-horizon agentic tasks, or multimodal settings is left for future work.

\section*{Ethics Statement}
This work only uses publicly available datasets and open-source or commercially available LLMs, all of which are properly cited and used in accordance with their licenses and terms of use. Our method does not involve human subjects or the collection of private data, and we foresee no direct negative societal impacts beyond those generally associated with LLM-based multi-agent systems. We use LLMs only for grammar checking and polishing of our manually written content.

\bibliography{reference}

@misc{langgraph,
  author = {LangChain Inc.},
  title = {LangGraph},
  year = {2024},
  howpublished = {\url{https://github.com/langchain-ai/langgraph}}
}

@misc{autogpt,
  author       = {{Significant Gravitas}},
  title        = {AutoGPT},
  year         = {2023},
  howpublished = {\url{https://github.com/Significant-Gravitas/AutoGPT}}
}

@inproceedings{wu2024autogen,
  title={AutoGen: Enabling Next-Gen LLM Applications via Multi-Agent Conversation},
  author={Wu, Qingyun and Bansal, Gagan and Zhang, Jieyu and Wu, Yiran and Li, Beibin and Zhu, Erkang and Jiang, Li and Zhang, Xiaoyun and Zhang, Shaokun and Liu, Jiale and others},
  booktitle={ICLR 2024 Workshop on Large Language Model (LLM) Agents}
}

@article{li2025search,
  title={Search-o1: Agentic search-enhanced large reasoning models},
  author={Li, Xiaoxi and Dong, Guanting and Jin, Jiajie and Zhang, Yuyao and Zhou, Yujia and Zhu, Yutao and Zhang, Peitian and Dou, Zhicheng},
  journal={arXiv preprint arXiv:2501.05366},
  year={2025}
}

@article{zheng2025deepresearcher,
  title={Deepresearcher: Scaling deep research via reinforcement learning in real-world environments},
  author={Zheng, Yuxiang and Fu, Dayuan and Hu, Xiangkun and Cai, Xiaojie and Ye, Lyumanshan and Lu, Pengrui and Liu, Pengfei},
  journal={arXiv preprint arXiv:2504.03160},
  year={2025}
}

@inproceedings{zheng2023codegeex,
  title={Codegeex: A pre-trained model for code generation with multilingual benchmarking on humaneval-x},
  author={Zheng, Qinkai and Xia, Xiao and Zou, Xu and Dong, Yuxiao and Wang, Shan and Xue, Yufei and Shen, Lei and Wang, Zihan and Wang, Andi and Li, Yang and others},
  booktitle={Proceedings of the 29th ACM SIGKDD Conference on Knowledge Discovery and Data Mining},
  pages={5673--5684},
  year={2023}
}

@article{jin2023surrealdriver,
  title={Surrealdriver: Designing generative driver agent simulation framework in urban contexts based on large language model},
  author={Jin, Ye and Shen, Xiaoxi and Peng, Huiling and Liu, Xiaoan and Qin, Jingli and Li, Jiayang and Xie, Jintao and Gao, Peizhong and Zhou, Guyue and Gong, Jiangtao},
  journal={arXiv preprint arXiv:2309.13193},
  volume={5},
  number={7},
  pages={8},
  year={2023}
}

@inproceedings{zhuge2024gptswarm,
  title={Gptswarm: Language agents as optimizable graphs},
  author={Zhuge, Mingchen and Wang, Wenyi and Kirsch, Louis and Faccio, Francesco and Khizbullin, Dmitrii and Schmidhuber, J{\"u}rgen},
  booktitle={Forty-first International Conference on Machine Learning},
  year={2024}
}

@inproceedings{GDesigner,
  title={G-Designer: Architecting Multi-agent Communication Topologies via Graph Neural Networks},
  author={Zhang, Guibin and Yue, Yanwei and Sun, Xiangguo and Wan, Guancheng and Yu, Miao and Fang, Junfeng and Wang, Kun and Chen, Tianlong and Cheng, Dawei},
  booktitle={Forty-second International Conference on Machine Learning}
}

@article{chen2021evaluating,
  title={Evaluating large language models trained on code},
  author={Chen, Mark},
  journal={arXiv preprint arXiv:2107.03374},
  year={2021}
}

@inproceedings{du2023improving,
  title={Improving factuality and reasoning in language models through multiagent debate},
  author={Du, Yilun and Li, Shuang and Torralba, Antonio and Tenenbaum, Joshua B and Mordatch, Igor},
  booktitle={Forty-first International Conference on Machine Learning},
  year={2023}
}

@inproceedings{wangself,
  title={Self-Consistency Improves Chain of Thought Reasoning in Language Models},
  author={Wang, Xuezhi and Wei, Jason and Schuurmans, Dale and Le, Quoc V and Chi, Ed H and Narang, Sharan and Chowdhery, Aakanksha and Zhou, Denny},
  booktitle={The Eleventh International Conference on Learning Representations},
  year={2023}
}

@article{fansi2022ddxplus,
  title={Ddxplus: A new dataset for automatic medical diagnosis},
  author={Fansi Tchango, Arsene and Goel, Rishab and Wen, Zhi and Martel, Julien and Ghosn, Joumana},
  journal={Advances in neural information processing systems},
  volume={35},
  pages={31306--31318},
  year={2022}
}

@inproceedings{yang2018hotpotqa,
  title={HotpotQA: A dataset for diverse, explainable multi-hop question answering},
  author={Yang, Zhilin and Qi, Peng and Zhang, Saizheng and Bengio, Yoshua and Cohen, William and Salakhutdinov, Ruslan and Manning, Christopher D},
  booktitle={Proceedings of the 2018 conference on empirical methods in natural language processing},
  pages={2369--2380},
  year={2018}
}

@inproceedings{lai2023ds,
  title={DS-1000: A natural and reliable benchmark for data science code generation},
  author={Lai, Yuhang and Li, Chengxi and Wang, Yiming and Zhang, Tianyi and Zhong, Ruiqi and Zettlemoyer, Luke and Yih, Wen-tau and Fried, Daniel and Wang, Sida and Yu, Tao},
  booktitle={International Conference on Machine Learning},
  pages={18319--18345},
  year={2023},
  organization={PMLR}
}

@inproceedings{ling2017program,
  title={Program Induction by Rationale Generation: Learning to Solve and Explain Algebraic Word Problems},
  author={Ling, Wang and Yogatama, Dani and Dyer, Chris and Blunsom, Phil},
  booktitle={Proceedings of the 55th Annual Meeting of the Association for Computational Linguistics (Volume 1: Long Papers)},
  pages={158--167},
  year={2017}
}

@article{patel2021nlp,
  title={Are NLP models really able to solve simple math word problems?},
  author={Patel, Arkil and Bhattamishra, Satwik and Goyal, Navin},
  journal={arXiv preprint arXiv:2103.07191},
  year={2021}
}

@inproceedings{roy2015solving,
  title={Solving general arithmetic word problems},
  author={Roy, Subhro and Roth, Dan},
  booktitle={Proceedings of the 2015 conference on empirical methods in natural language processing},
  pages={1743--1752},
  year={2015}
}

@article{cobbe2021training,
  title={Training verifiers to solve math word problems},
  author={Cobbe, Karl and Kosaraju, Vineet and Bavarian, Mohammad and Chen, Mark and Jun, Heewoo and Kaiser, Lukasz and Plappert, Matthias and Tworek, Jerry and Hilton, Jacob and Nakano, Reiichiro and others},
  journal={arXiv preprint arXiv:2110.14168},
  year={2021}
}

@inproceedings{hendrycksmeasuring,
  title={Measuring Massive Multitask Language Understanding},
  author={Hendrycks, Dan and Burns, Collin and Basart, Steven and Zou, Andy and Mazeika, Mantas and Song, Dawn and Steinhardt, Jacob},
  booktitle={International Conference on Learning Representations}
}

@article{mackay1992practical,
  title={A practical Bayesian framework for backpropagation networks},
  author={MacKay, David JC},
  journal={Neural computation},
  volume={4},
  number={3},
  pages={448--472},
  year={1992},
  publisher={MIT Press One Rogers Street, Cambridge, MA 02142-1209, USA journals-info~…}
}

@article{reimers1908sentence,
  title={Sentence embeddings using siamese BERT-networks. arXiv 2019},
  author={Reimers, Nils and Gurevych, I Sentence-BERT and others},
  journal={arXiv preprint arXiv:1908.10084},
  volume={10},
  year={1908}
}

@inproceedings{cho2014learning,
  title={Learning phrase representations using RNN encoder--decoder for statistical machine translation},
  author={Cho, Kyunghyun and Van Merri{\"e}nboer, Bart and Gul{\c{c}}ehre, {\c{C}}a{\u{g}}lar and Bahdanau, Dzmitry and Bougares, Fethi and Schwenk, Holger and Bengio, Yoshua},
  booktitle={Proceedings of the 2014 conference on empirical methods in natural language processing (EMNLP)},
  pages={1724--1734},
  year={2014}
}

@inproceedings{burges2005learning,
  title={Learning to rank using gradient descent},
  author={Burges, Chris and Shaked, Tal and Renshaw, Erin and Lazier, Ari and Deeds, Matt and Hamilton, Nicole and Hullender, Greg},
  booktitle={Proceedings of the 22nd international conference on Machine learning},
  pages={89--96},
  year={2005}
}

@article{aepo,
  title={Agentic entropy-balanced policy optimization},
  author={Dong, Guanting and Bao, Licheng and Wang, Zhongyuan and Zhao, Kangzhi and Li, Xiaoxi and Jin, Jiajie and Yang, Jinghan and Mao, Hangyu and Zhang, Fuzheng and Gai, Kun and others},
  journal={arXiv preprint arXiv:2510.14545},
  year={2025}
}

@article{arpo,
  title={Agentic reinforced policy optimization},
  author={Dong, Guanting and Mao, Hangyu and Ma, Kai and Bao, Licheng and Chen, Yifei and Wang, Zhongyuan and Chen, Zhongxia and Du, Jiazhen and Wang, Huiyang and Zhang, Fuzheng and others},
  journal={arXiv preprint arXiv:2507.19849},
  year={2025}
}

@article{pesce2023learning,
  title={Learning multi-agent coordination through connectivity-driven communication},
  author={Pesce, Emanuele and Montana, Giovanni},
  journal={Machine Learning},
  volume={112},
  number={2},
  pages={483--514},
  year={2023},
  publisher={Springer}
}

@inproceedings{zhangaflow,
  title={AFlow: Automating Agentic Workflow Generation},
  author={Zhang, Jiayi and Xiang, Jinyu and Yu, Zhaoyang and Teng, Fengwei and Chen, Xiong-Hui and Chen, Jiaqi and Zhuge, Mingchen and Cheng, Xin and Hong, Sirui and Wang, Jinlin and others},
  booktitle={The Thirteenth International Conference on Learning Representations},
  year={2024}
}

@article{li2025assemble,
  title={Assemble your crew: Automatic multi-agent communication topology design via autoregressive graph generation},
  author={Li, Shiyuan and Liu, Yixin and Wen, Qingsong and Zhang, Chengqi and Pan, Shirui},
  journal={arXiv preprint arXiv:2507.18224},
  year={2025}
}

@article{damani2024learning,
  title={Learning how hard to think: Input-adaptive allocation of lm computation},
  author={Damani, Mehul and Shenfeld, Idan and Peng, Andi and Bobu, Andreea and Andreas, Jacob},
  journal={arXiv preprint arXiv:2410.04707},
  year={2024}
}

@article{lee2025confidence,
  title={Confidence-Guided Stepwise Model Routing for Cost-Efficient Reasoning},
  author={Lee, Sangmook and Kim, Dohyung and Koh, Hyukhun and Yang, Nakyeong and Jung, Kyomin},
  journal={arXiv preprint arXiv:2511.06190},
  year={2025}
}

@article{wei2022chain,
  title={Chain-of-thought prompting elicits reasoning in large language models},
  author={Wei, Jason and Wang, Xuezhi and Schuurmans, Dale and Bosma, Maarten and Xia, Fei and Chi, Ed and Le, Quoc V and Zhou, Denny and others},
  journal={Advances in neural information processing systems},
  volume={35},
  pages={24824--24837},
  year={2022}
}

@inproceedings{hong2023metagpt,
  title={MetaGPT: Meta programming for a multi-agent collaborative framework},
  author={Hong, Sirui and Zhuge, Mingchen and Chen, Jonathan and Zheng, Xiawu and Cheng, Yuheng and Wang, Jinlin and Zhang, Ceyao and Wang, Zili and Yau, Steven Ka Shing and Lin, Zijuan and others},
  booktitle={The twelfth international conference on learning representations},
  year={2023}
}

@article{yao2023tree,
  title={Tree of thoughts: Deliberate problem solving with large language models, 2023},
  author={Yao, Shunyu and Yu, Dian and Zhao, Jeffrey and Shafran, Izhak and Griffiths, Thomas L and Cao, Yuan and Narasimhan, Karthik},
  journal={URL https://arxiv. org/abs/2305.10601},
  volume={3},
  pages={1},
  year={2023}
}

@article{zhang2025safesieve,
  title={SafeSieve: From Heuristics to Experience in Progressive Pruning for LLM-based Multi-Agent Communication},
  author={Zhang, Ruijia and Zhao, Xinyan and Wang, Ruixiang and Chen, Sigen and Zhang, Guibin and Zhang, An and Wang, Kun and Wen, Qingsong},
  journal={arXiv preprint arXiv:2508.11733},
  year={2025}
}

@inproceedings{wang2025evoagentx,
  title={Evoagentx: An automated framework for evolving agentic workflows},
  author={Wang, Yingxu and Liu, Siwei and Fang, Jinyuan and Meng, Zaiqiao},
  booktitle={Proceedings of the 2025 Conference on Empirical Methods in Natural Language Processing: System Demonstrations},
  pages={643--655},
  year={2025}
}

@article{liu2023dynamic,
  title={Dynamic llm-agent network: An llm-agent collaboration framework with agent team optimization},
  author={Liu, Zijun and Zhang, Yanzhe and Li, Peng and Liu, Yang and Yang, Diyi},
  journal={arXiv preprint arXiv:2310.02170},
  year={2023}
}

@misc{tora,
      title={ToRA: A Tool-Integrated Reasoning Agent for Mathematical Problem Solving}, 
      author={Zhibin Gou and Zhihong Shao and Yeyun Gong and Yelong Shen and Yujiu Yang and Minlie Huang and Nan Duan and Weizhu Chen},
      year={2024},
      eprint={2309.17452},
      archivePrefix={arXiv},
      primaryClass={cs.CL},
}

@inproceedings{DBLP:conf/iclr/ShazeerMMDLHD17,
  author       = {Noam Shazeer and
                  Azalia Mirhoseini and
                  Krzysztof Maziarz and
                  Andy Davis and
                  Quoc V. Le and
                  Geoffrey E. Hinton and
                  Jeff Dean},
  title        = {Outrageously Large Neural Networks: The Sparsely-Gated Mixture-of-Experts
                  Layer},
  booktitle    = {5th International Conference on Learning Representations, {ICLR} 2017,
                  Toulon, France, April 24-26, 2017, Conference Track Proceedings},
  publisher    = {OpenReview.net},
  year         = {2017},
  url          = {https://openreview.net/forum?id=B1ckMDqlg},
  bibsource    = {dblp computer science bibliography, https://dblp.org}
}

@inproceedings{xia2008listwise,
  title={Listwise approach to learning to rank: theory and algorithm},
  author={Xia, Fen and Liu, Tie-Yan and Wang, Jue and Zhang, Wensheng and Li, Hang},
  booktitle={Proceedings of the 25th international conference on Machine learning},
  pages={1192--1199},
  year={2008}
}

@inproceedings{schroff2015facenet,
  title={Facenet: A unified embedding for face recognition and clustering},
  author={Schroff, Florian and Kalenichenko, Dmitry and Philbin, James},
  booktitle={Proceedings of the IEEE conference on computer vision and pattern recognition},
  pages={815--823},
  year={2015}
}

@article{zhou2022mixture,
  title={Mixture-of-experts with expert choice routing},
  author={Zhou, Yanqi and Lei, Tao and Liu, Hanxiao and Du, Nan and Huang, Yanping and Zhao, Vincent and Dai, Andrew M and Le, Quoc V and Laudon, James and others},
  journal={Advances in Neural Information Processing Systems},
  volume={35},
  pages={7103--7114},
  year={2022}
}

@inproceedings{DBLP:conf/iclr/LepikhinLXCFHKS21,
  author       = {Dmitry Lepikhin and
                  HyoukJoong Lee and
                  Yuanzhong Xu and
                  Dehao Chen and
                  Orhan Firat and
                  Yanping Huang and
                  Maxim Krikun and
                  Noam Shazeer and
                  Zhifeng Chen},
  title        = {GShard: Scaling Giant Models with Conditional Computation and Automatic
                  Sharding},
  booktitle    = {9th International Conference on Learning Representations, {ICLR} 2021,
                  Virtual Event, Austria, May 3-7, 2021},
  publisher    = {OpenReview.net},
  year         = {2021},
  url          = {https://openreview.net/forum?id=qrwe7XHTmYb},
  bibsource    = {dblp computer science bibliography, https://dblp.org}
}

@inproceedings{DBLP:conf/acl/Xie0SCZWZ25,
  author       = {Zhentao Xie and
                  Chengcheng Han and
                  Jinxin Shi and
                  Wenjun Cui and
                  Xin Zhao and
                  Xingjiao Wu and
                  Jiabao Zhao},
  editor       = {Wanxiang Che and
                  Joyce Nabende and
                  Ekaterina Shutova and
                  Mohammad Taher Pilehvar},
  title        = {RMoA: Optimizing Mixture-of-Agents through Diversity Maximization
                  and Residual Compensation},
  booktitle    = {Findings of the Association for Computational Linguistics, {ACL} 2025,
                  Vienna, Austria, July 27 - August 1, 2025},
  series       = {Findings of {ACL}},
  pages        = {6575--6602},
  publisher    = {Association for Computational Linguistics},
  year         = {2025},
  url          = {https://aclanthology.org/2025.findings-acl.342/},
  bibsource    = {dblp computer science bibliography, https://dblp.org}
}

@inproceedings{wangmixture,
  title={Mixture-of-Agents Enhances Large Language Model Capabilities},
  author={Wang, Junlin and Jue, WANG and Athiwaratkun, Ben and Zhang, Ce and Zou, James},
  booktitle={The Thirteenth International Conference on Learning Representations},
  year={2025}
}

@inproceedings{liu2022temporal,
  title={Temporal dynamic weighted graph convolution for multi-agent reinforcement learning},
  author={Liu, Yuntao and Dou, Yong and Li, Yuan and Xu, Xinhai and Liu, Donghong},
  booktitle={Proceedings of the Annual Meeting of the Cognitive Science Society},
  volume={44},
  number={44},
  year={2022}
}

@article{chan2023chateval,
  title={Chateval: Towards better llm-based evaluators through multi-agent debate},
  author={Chan, Chi-Min and Chen, Weize and Su, Yusheng and Yu, Jianxuan and Xue, Wei and Zhang, Shanghang and Fu, Jie and Liu, Zhiyuan},
  journal={arXiv preprint arXiv:2308.07201},
  year={2023}
}

@article{hu2024learning,
  title={Learning multi-agent communication from graph modeling perspective},
  author={Hu, Shengchao and Shen, Li and Zhang, Ya and Tao, Dacheng},
  journal={arXiv preprint arXiv:2405.08550},
  year={2024}
}

@article{qian2024scaling,
  title={Scaling large language model-based multi-agent collaboration},
  author={Qian, Chen and Xie, Zihao and Wang, Yifei and Liu, Wei and Zhu, Kunlun and Xia, Hanchen and Dang, Yufan and Du, Zhuoyun and Chen, Weize and Yang, Cheng and others},
  journal={arXiv preprint arXiv:2406.07155},
  year={2024}
}

\clearpage
\appendix
\appendix

% Hide the main-body entries, and re-enable the depth from this point on,
% so the box below only lists the appendix.
\setcounter{tocdepth}{-1}
\addtocontents{toc}{\protect\setcounter{tocdepth}{2}}
\begin{tcolorbox}[colback=gray!3, colframe=gray!50, title={\large\bfseries\sffamily Table of Contents}, coltitle=black, toptitle=4pt, bottomtitle=4pt, left=8pt, right=8pt, top=8pt, bottom=8pt, arc=2pt]
\renewcommand{\contentsname}{}
\vspace{-1.5em}
\tableofcontents
\end{tcolorbox}

\section{Dataset Details}
\label{app:dataset_details}

We evaluate \ourmethod on nine benchmarks spanning diverse reasoning settings.
Specifically, MMLU is used for broad-domain reasoning; GSM8K, MultiArith, SVAMP, and AQuA for mathematical reasoning; HumanEval and DS-1000 for coding and program understanding; HotpotQA for multi-hop web-based question answering; and DDXPlus for medical diagnosis reasoning.
This benchmark suite follows the same overall task coverage described in the main paper and is also consistent with the uploaded reference paper ST-EVO, which uses the same nine-benchmark evaluation protocol.

For all datasets, we follow the standard public splits and official evaluation protocols whenever available.
To ensure fair comparison across methods, all baselines and \ourmethod are instantiated with the same backbone model, \texttt{gpt-oss-120b}, and are evaluated on the same benchmark-specific query sets.
For optimizable methods, including \ourmethod, we adopt a unified few-shot adaptation protocol: for each benchmark, we sample 40--80 queries from the corresponding training split to optimize the routing mechanism while keeping the underlying LLM agents frozen.

\begin{table}[h]
\centering
\small
\resizebox{\linewidth}{!}{%
\begin{tabular}{l l l}
\toprule
Benchmark & Task Type & Metric \\
\midrule
MMLU & Broad-domain reasoning & Accuracy \\
GSM8K & Grade-school math reasoning & Accuracy \\
MultiArith & Arithmetic reasoning & Accuracy \\
SVAMP & Mathematical word problems & Accuracy \\
AQuA & Algebraic reasoning & Accuracy \\
HumanEval & Code generation & Pass@1 / Accuracy \\
DS-1000 & Code reasoning and debugging & Pass@1 / Accuracy \\
HotpotQA & Multi-hop question answering & Accuracy / F1 \\
DDXPlus & Medical diagnosis reasoning & Accuracy \\
\bottomrule
\end{tabular}}
\caption{Benchmarks used in \ourmethod.}
\label{tab:DMoA_datasets}
\end{table}

\subsection{Method Details}
\subsection{Task-specific Agent Pool Configuration}
\label{app:agent_pool}

In \ourmethod, each agent is instantiated as
\[
a_i = \{\mathrm{LLM}_i, \mathrm{Profile}_i, \mathrm{Tool}_i\},
\]
where the agent profile is realized through system prompts, and the tool set specifies its executable external functionalities.
This design follows the formulation in the main paper, where routed agents can differ in their specialized capabilities and are dynamically selected at each reasoning step.

Following the provided implementation, the practical agent pool is organized around three task-oriented groups: code-oriented agents, math-oriented agents, and open-domain QA agents.
For readability, we summarize tools by their functional names rather than raw function names from the code.

\paragraph{Code-oriented agents.}
These agents focus on software design, implementation, testing, and repair.
Representative roles include \emph{Project Manager}, \emph{Algorithm Designer}, \emph{Programming Expert}, \emph{Test Analyst}, and \emph{Bug Fixer}.
Their summarized tool categories include \emph{Code Execution}, \emph{Unit Testing}, \emph{Script Running}, and optionally \emph{Knowledge Lookup}.

\paragraph{Math-oriented agents.}
These agents focus on mathematical derivation, symbolic reasoning, numerical verification, and consistency checking.
Representative roles include \emph{Math Solver}, \emph{Mathematical Analyst}, \emph{Programming Expert}, and \emph{Inspector}.
Their summarized tool categories include \emph{Calculator} and \emph{Code Execution}.

\paragraph{Open-domain QA agents.}
These agents target knowledge-intensive reasoning and domain-specific interpretation.
Representative roles include \emph{Knowledgeable Expert}, \emph{Wiki Searcher}, \emph{Critic}, \emph{Mathematician}, \emph{Psychologist}, \emph{Historian}, \emph{Doctor}, \emph{Lawyer}, \emph{Economist}, and \emph{Programmer}.
Their summarized tool categories include \emph{Encyclopedic Search}, \emph{Domain Knowledge Retrieval}, \emph{Calculator}, and \emph{Code Execution}.

\begin{table*}[t]
\centering
\small
\begin{tabular}{p{2.1cm} p{6.5cm} p{4.0cm}}
\toprule
Group & Representative Roles & Tool Categories \\
\midrule
Code &
Project Manager; Algorithm Designer; Programming Expert; Test Analyst; Bug Fixer &
Code Execution; Unit Testing; Script Running; Knowledge Lookup \\
Math &
Math Solver; Mathematical Analyst; Programming Expert; Inspector &
Calculator; Code Execution \\
Open-domain QA &
Knowledgeable Expert; Wiki Searcher; Critic; Mathematician; Psychologist; Historian; Doctor; Lawyer; Economist; Programmer &
Encyclopedic Search; Domain Knowledge Retrieval; Calculator; Code Execution \\
\bottomrule
\end{tabular}
\caption{Task-oriented role families and summarized tool categories in the agent pool.}
\label{tab:agent_pool_summary}
\end{table*}

\paragraph{Summarizer agent.}
In addition to task-specific experts, \ourmethod introduces a summarizer agent to aggregate intermediate dialogue trajectories and produce the final output.
As described in the main paper, the summarizer can either generate the final response when the maximum reasoning budget is reached or autonomously determine whether the current reasoning process should terminate earlier.

\subsection{Prompt Construction}
\label{app:prompts}

For each routed agent, the input prompt consists of two major parts:
\[
\mathcal{P}_{i,j} = (\mathcal{P}_{\mathrm{sys}}, \mathcal{P}_{\mathrm{last}}).
\]
Here, the system prompt
\[
\mathcal{P}_{\mathrm{sys}} = (\mathrm{Profile}_{i,j}, \mathcal{P}_{\mathrm{usr}})
\]
contains the role profile of the routed agent together with the original user query, while
\[
\mathcal{P}_{\mathrm{last}}
\]
contains the collected responses from the previous reasoning step.

\paragraph{Generic agent system prompt.}
A generic template is shown below:

\begin{quote}
\small
\textbf{[System Prompt]}\\
You are a specialized expert for solving the given task. Your role is: \texttt{<ROLE>}.\\
You should reason carefully based on the user query and the intermediate responses from other agents.
When useful, you may invoke the following tools: \texttt{<TOOLS>}.\\
Your goal is to provide a concise but high-quality intermediate solution that is maximally helpful for subsequent agents.
\end{quote}

\paragraph{Context aggregation prompt.}
Given the responses from the previous reasoning step
\[
\{\mathcal{R}_{i-1,1}, \ldots, \mathcal{R}_{i-1,\mathrm{k}_{i-1}}\},
\]
the contextual prompt is constructed as:

\begin{quote}
\small
\textbf{[Context Prompt]}\\
You are given the original user query and a set of intermediate responses generated by other agents.
Please read them carefully, identify the most reliable useful information, and continue the reasoning
from the current stage. Avoid repeating previous content unless it is necessary for correction or synthesis.

Responses from previous-step agents:
\begin{enumerate}
    \item \texttt{<Response 1>}
    \item \texttt{<Response 2>}
    \item \ldots
    \item \texttt{<Response k>}
\end{enumerate}
\end{quote}

\paragraph{Summarizer prompt.}
The summarizer receives the original query and all current-step responses:

\begin{quote}
\small
\textbf{[Summarizer Prompt]}\\
You are the final summarizer. Given the original user query and the candidate responses from multiple agents,
your job is to synthesize them into a single final answer. If the current evidence is already sufficient to answer
the question reliably, output the final answer directly. Otherwise, explicitly indicate that more reasoning is needed.
\end{quote}

\paragraph{Task-specific prompt instantiation.}
For different benchmarks, we instantiate task-specific role prompts:
(i) mathematical analyst roles for GSM8K, MultiArith, SVAMP, and AQuA;
(ii) code analyst, programmer, tester, and bug-fixing roles for HumanEval and DS-1000;
(iii) evidence-grounded search and critique roles for HotpotQA and broad-domain QA;
(iv) diagnosis- and domain-specific roles for DDXPlus and general-domain knowledge tasks.

\section{Training Details}
\subsection{Training Algorithm}
\label{app:train_algo}

Algorithm~\ref{alg:DMoA_train} summarizes the optimization-time pipeline of \ourmethod.

\begin{algorithm}[h]
\caption{Training of \ourmethod with entropy supervision}
\label{alg:DMoA_train}
\small
\begin{algorithmic}[1]
\STATE Initialize the Sentence Transformer encoder $\texttt{ST}$, the RNN-Router $\texttt{RNN}$, and the linear routing head.
\FOR{each training query $\mathcal{P}_{usr}$}
    \STATE Compute the initial semantic vector $\mathrm{X}_1 = \texttt{ST}(\mathcal{P}_{usr})$.
    \STATE Initialize hidden state $\mathrm{h}_0 = \mathbf{0}$.
    \FOR{$i = 1$ to $L_{\max}$}
        \STATE Compute routing logits $\mathrm{z}_i$ using the router.
        \STATE Execute all $\mathrm{N}$ candidate agents to collect predictive entropy vector $\mathcal{E}_i \in \mathbb{R}^{\mathrm{N}}$.
        \STATE Convert entropy to confidence targets:
        \[
        \mathcal{C}_i = \texttt{Softmax}(-\mathcal{E}_i).
        \]
        \STATE Compute routing probabilities:
        \[
        \mathcal{Z}_i = \texttt{Softmax}(\mathrm{z}_i).
        \]
        \STATE Compute the pair-wise ranking loss $\mathcal{L}_{\text{rank}}^{(i)}$.
        \STATE Aggregate the responses of the routed agents to form $\mathrm{X}_{i+1}$.
    \ENDFOR
    \STATE Optimize the router by minimizing
    \[
    \mathcal{L}_{\text{rank}} = \frac{1}{L}\sum_{i=1}^{L}\mathcal{L}_{\text{rank}}^{(i)}.
    \]
\ENDFOR
\end{algorithmic}
\end{algorithm}

\subsection{Inference Algorithm}
\label{app:infer_algo}

During inference, \ourmethod performs sparse activation and only executes the routed agents selected at each reasoning step.
Algorithm~\ref{alg:DMoA_infer} summarizes the sparse inference procedure.

\begin{algorithm}[h]
\caption{Sparse inference of \ourmethod}
\label{alg:DMoA_infer}
\small
\begin{algorithmic}[1]
\STATE Input query $\mathcal{P}_{usr}$.
\STATE Compute $\mathrm{X}_1 = \texttt{ST}(\mathcal{P}_{usr})$ and initialize hidden state $\mathrm{h}_0$.
\FOR{$i = 1$ to $L_{\max}$}
    \STATE Compute routing logits $\mathrm{z}_i$.
    \STATE Determine the number of routed agents:
    \[
    \mathrm{k}_i = \texttt{Min}\left(\mathrm{K}, \texttt{Count}\left(\texttt{Softmax}(\mathrm{z}_i / \tau) \ge \frac{1}{\mathrm{N}}\right)\right).
    \]
    \STATE Select the top-$\mathrm{k}_i$ agents and execute them.
    \STATE Aggregate their responses to obtain $\mathrm{X}_{i+1}$.
    \STATE Invoke the summarizer to determine whether to terminate.
    \IF{termination criterion is satisfied}
        \STATE Output the summarized final answer and stop.
    \ENDIF
\ENDFOR
\STATE Output the final summarizer response at step $L_{\max}$.
\end{algorithmic}
\end{algorithm}

\subsection{Optimization Hyperparameters}
\label{app:hyper}

Unless otherwise stated, we use the following default hyperparameters for \ourmethod:
\begin{itemize}
    \item Sentence Transformer: \texttt{all-MiniLM-L6-v2}, embedding dimension $D = 384$;
    \item router backbone: GRU;
    \item optimizer: \texttt{AdamW};
    \item learning rate: \texttt{1e-3};
    \item batch size: \texttt{8};
    \item number of training epochs: \texttt{3};
    \item routing temperature $\tau$: \texttt{0.1};
    \item maximum reasoning steps $L_{\max}$: \texttt{20};
    \item gradient clipping norm: \texttt{1.0}.
\end{itemize}

We optimize only the routing-related modules unless otherwise noted.
The parameters of the underlying LLM agents remain frozen during training and inference-time adaptation.

\subsection{Few-shot Adaptation Protocol}
\label{app:fewshot}

For optimizable methods, including \ourmethod and prior adaptive baselines, we follow a unified few-shot adaptation protocol.
Specifically, for each benchmark, we sample 40--80 training queries as the adaptation set.
These queries are used only to optimize the topology- or routing-related modules, while the backbone LLM remains fixed.
All methods share the same backbone model and use the same tool budget whenever applicable.

\subsection{Test-Time Training Protocol}
\label{app:ttt}

\ourmethod naturally supports test-time training (TTT) because the predictive entropy of each candidate agent can be collected without external annotations.
In our TTT setting, for each benchmark, we first process the initial 10--30 test queries in a dense mode, where all $\mathrm{N}$ agents are executed in each reasoning step to collect entropy-based supervision signals.
We then update the router using the ranking loss and switch back to sparse activation for the remaining test queries.

Importantly, TTT does not use ground-truth labels from the test set.
The adaptation signal is entirely self-supervised and derived from the predictive entropy of the candidate agents.
This protocol matches the high-level description in the main paper, which states that \ourmethod collects predictive entropy on 10--30 test queries and optimizes the routing mechanism to adapt to the task during inference.

\subsection{Termination Strategy and Summarizer Details}
\label{app:termination}

We consider two stopping strategies:
\begin{enumerate}
    \item \textbf{Fixed-budget stopping}, where \ourmethod stops after a predefined maximum number of reasoning steps $L_{\max}$;
    \item \textbf{Summarizer-based stopping}, where a summarizer agent decides whether the current evidence is sufficient.
\end{enumerate}

In the second setting, the summarizer outputs either:
\begin{itemize}
    \item \texttt{[FINAL]} followed by the final answer, or
    \item \texttt{[CONTINUE]} if more reasoning is required.
\end{itemize}

This design allows \ourmethod to adaptively allocate reasoning depth according to task complexity.
It also matches the description in the main paper that the summarizer agent can autonomously determine at which step to terminate the reasoning process.

\subsection{Baseline Reproduction Details}
\label{app:baseline}

To ensure fair comparison, all baselines are instantiated using the same backbone model and, when possible, the same tool budget as \ourmethod.
For graph-based methods, we follow the original dialogue-iteration settings reported in their papers.
For methods requiring topology optimization, we adopt the same few-shot adaptation budget as \ourmethod.

In addition, for the efficiency comparison, we align the agent budget across methods.
For graph-based baselines, we set the number of participating agents to $\mathrm{N}_s$; for \ourmethod, we constrain the maximum routed-agent budget by setting $\mathrm{K} = \mathrm{N}_s$.
This follows the fairness protocol described in the main paper, where the same agent budget is imposed on all compared multi-agent systems.

\subsection{Scalability}
\label{app:scalability}

\begin{table}[!htbp]
\centering
\resizebox{0.88\linewidth}{!}{
\begin{tabular}{c|c|ccccc}
\toprule
Dataset & Metric & $(4,2)$ & $(6,3)$ & $(10,4)$ & $(15,6)$ & $(20,7)$ \\
\midrule
\multirow{2}{*}{MMLU}
& Accuracy (\%) & 87.3 & 89.4 & 91.2 & 92.3 & 92.5 \\
& \#Tokens & 536,428 & 748,913 & 1,032,684 & 1,431,752 & 1,816,395 \\
\midrule
\multirow{2}{*}{GSM8K}
& Accuracy (\%) & 92.1 & 94.6 & 97.4 & 98.9 & 99.2 \\
& \#Tokens & 742,615 & 1,036,284 & 1,426,917 & 1,982,436 & 2,514,872 \\
\midrule
\multirow{2}{*}{HumanEval}
& Accuracy (\%) & 84.1 & 90.2 & 93.6 & 95.6 & 96.2 \\
& \#Tokens & 147,382 & 204,756 & 276,913 & 386,428 & 492,175 \\
\midrule
\multirow{2}{*}{SVAMP}
& Accuracy (\%) & 89.8 & 92.1 & 94.3 & 94.8 & 95.2 \\
& \#Tokens & 381,946 & 528,314 & 732,681 & 1,018,925 & 1,293,476 \\
\bottomrule
\end{tabular}}
\caption{
Scalability of \ourmethod under different agent configurations on MMLU, GSM8K, HumanEval, and SVAMP.
The configuration $(\mathrm{N},\mathrm{K})$ denotes using $\mathrm{N}$ candidate agents and selecting $\mathrm{K}$ agents at each reasoning step.
As the configuration size increases, \ourmethod consistently improves accuracy across all benchmarks, while the marginal gain gradually saturates.
Token consumption grows near-linearly with the routing number $\mathrm{K}$.
}
\label{tab:scalability_nk_appendix}
\end{table}

\begin{figure}[!htbp]
    \centering
    \includegraphics[width=\columnwidth]{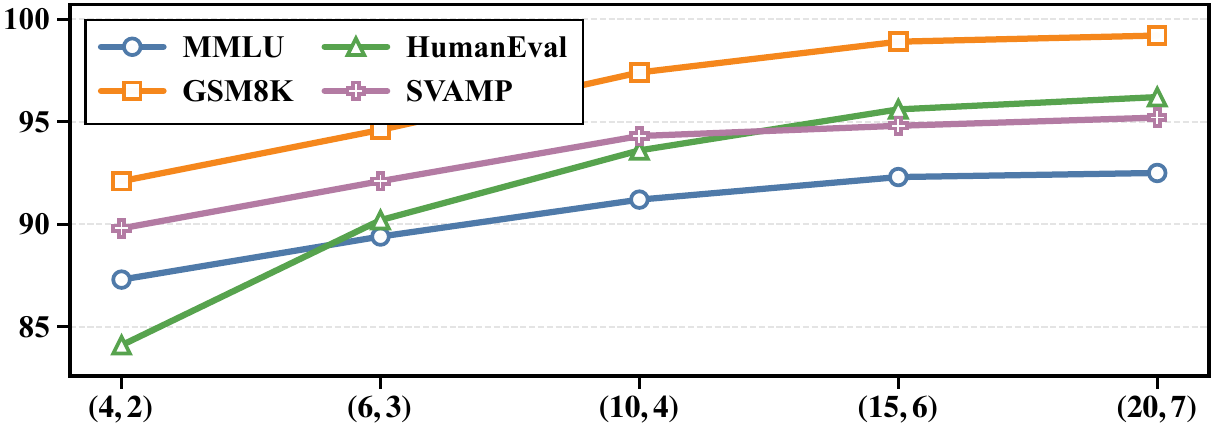}
    \caption{Scalability analyses on different routing configurations ($\mathrm{N}$, $\mathrm{K}$) in \ourmethod, across MMLU, HumanEval, GSM8K, and SVAMP.}
    \label{fig:scalability}
\end{figure}

Figure~\ref{fig:scalability} visualizes the same configurations, where the accuracies on all four benchmarks grow monotonically with $(\mathrm{N}, \mathrm{K})$.

\subsection{Robustness analysis under adversarial agents}
We evaluate the robustness of different multi-agent systems under prompt-based adversarial perturbations. Our setting follows the adversarial-agent protocol introduced in GPTSwarm~\citep{zhuge2024gptswarm}, where the attack is not applied by modifying the user query, but by replacing a subset of normal agents with adversarial agents. In the original GPTSwarm implementation, a normal agent is prompted to answer a multiple-choice MMLU question truthfully, while an adversarial agent is instructed with an internal system prompt such as:
\begin{quote}
\small
\textit{Answer a lie to the following question: \{question\}.}
\end{quote}

Following this evaluation paradigm, G-Designer~\citep{GDesigner} adopts a simplified robustness setting, where a system prompt attack is injected into one of five agents and the MMLU accuracy before and after the attack is compared. In our experiments, we extend this setting to different attack intensities by varying the proportion of adversarial agents in the system. Specifically, for a given attack ratio $\rho \in \{20\%, 40\%, 60\%, 80\%\}$, we randomly replace the corresponding fraction of agents with adversarial agents using the above adversarial prompt, while the remaining agents retain their normal task-solving prompts. We then run the complete multi-agent reasoning process without modifying the input questions or the aggregation protocol, and report the average accuracy before and after the attack across all evaluated benchmarks. This protocol directly measures whether a multi-agent system can maintain stable reasoning performance when a subset of its internal agents is intentionally corrupted.

\begin{figure*}[t]
    \centering

    \begin{minipage}[t]{0.48\textwidth}
        \centering
        \includegraphics[width=\linewidth]{Figures/robustness.pdf}
        
        \vspace{1mm}
        \small (a) $20\%$ attacked agents
    \end{minipage}
    \hfill
    \begin{minipage}[t]{0.48\textwidth}
        \centering
        \includegraphics[width=\linewidth]{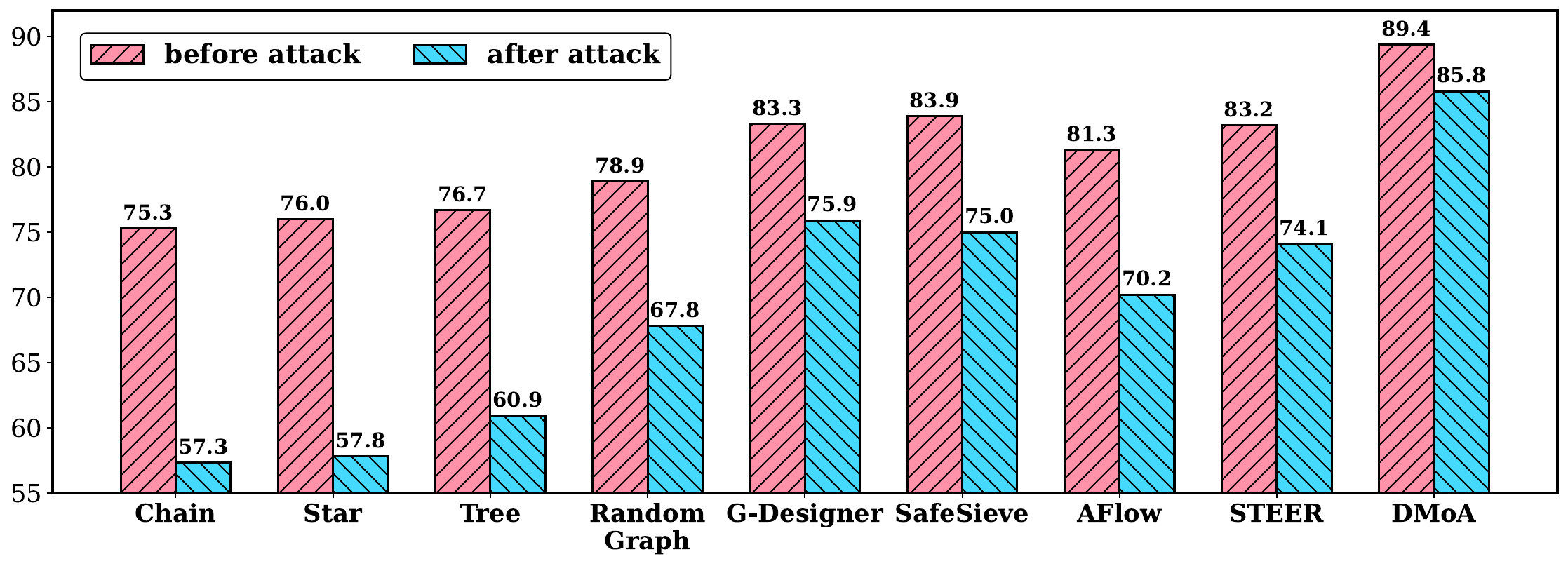}
        
        \vspace{1mm}
        \small (b) $40\%$ attacked agents
    \end{minipage}

    \vspace{2mm}

    \begin{minipage}[t]{0.48\textwidth}
        \centering
        \includegraphics[width=\linewidth]{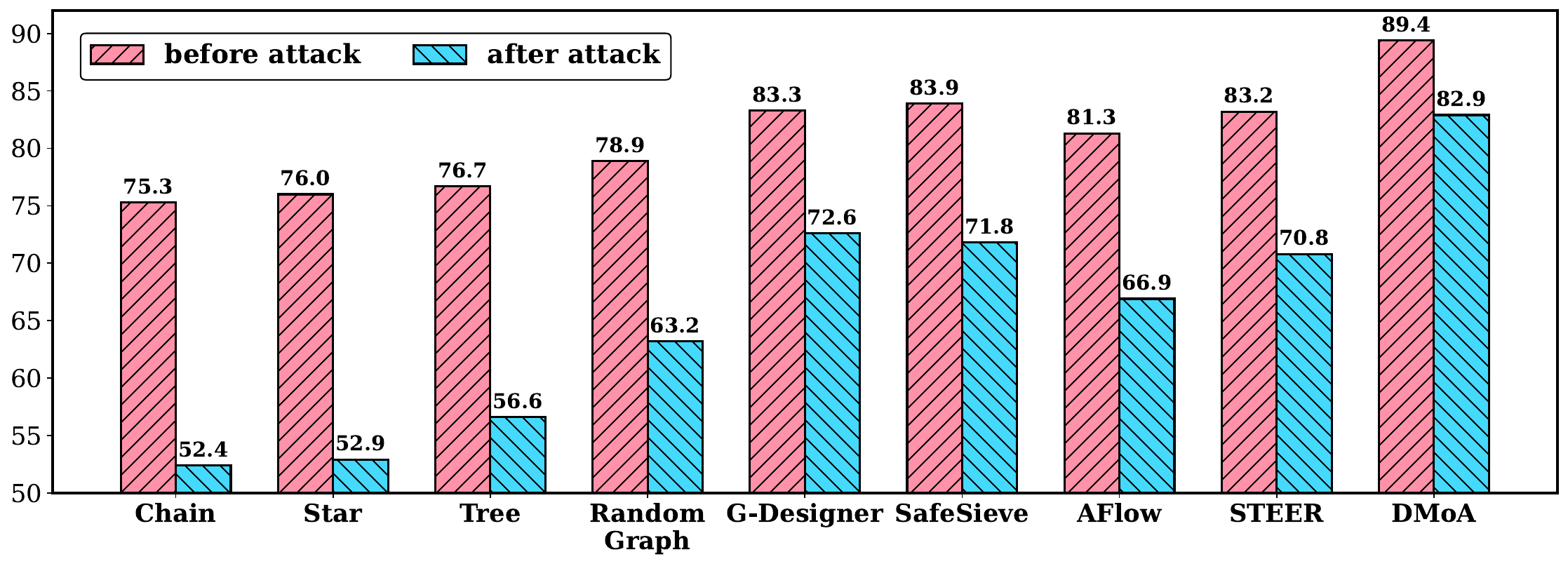}
        
        \vspace{1mm}
        \small (c) $60\%$ attacked agents
    \end{minipage}
    \hfill
    \begin{minipage}[t]{0.48\textwidth}
        \centering
        \includegraphics[width=\linewidth]{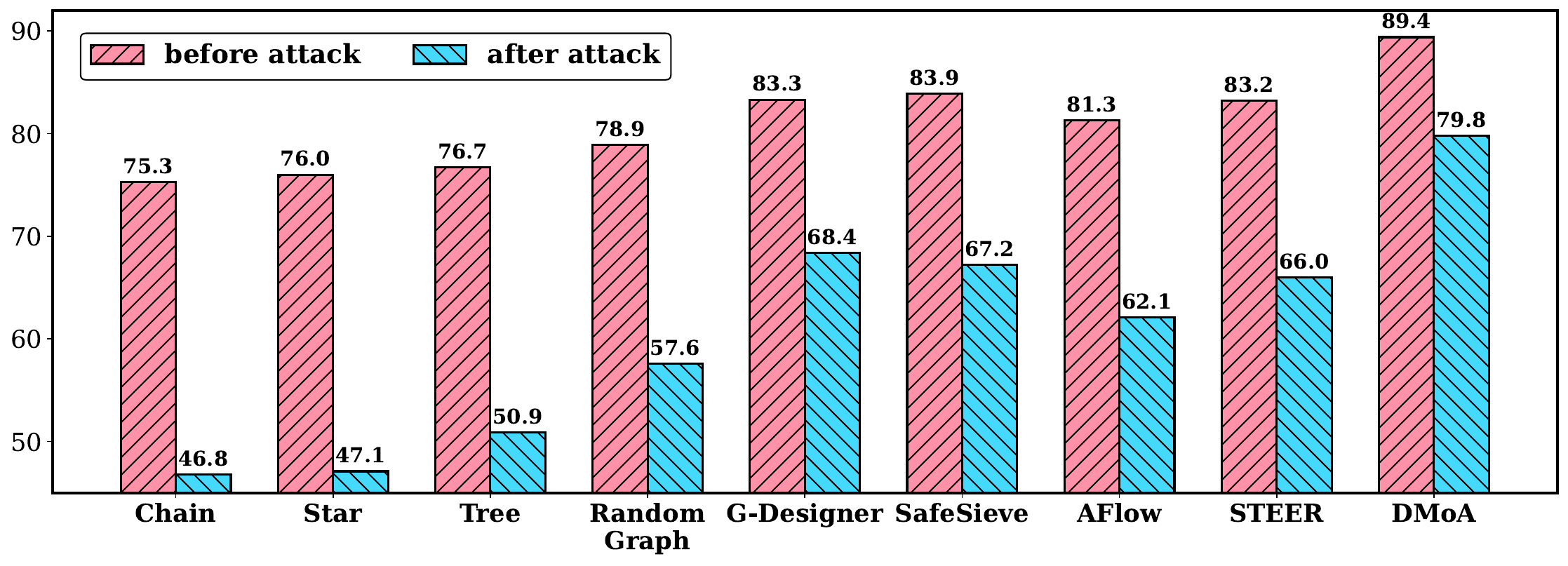}
        
        \vspace{1mm}
        \small (d) $80\%$ attacked agents
    \end{minipage}

    \caption{
    Robustness analysis under different adversarial-agent ratios.
    We compare the average accuracy of different multi-agent systems before and after prompt-based attacks, where a certain proportion of normal agents is replaced by adversarial agents.
    Results are reported under $20\%$, $40\%$, $60\%$, and $80\%$ attacked-agent settings.
    }
    \label{fig:robustness-all}
\end{figure*}

\section{Additional Experiments}
\label{app:added}

% ---------------------------------------------------------------------
\subsection{Agentic Coding and CLI Benchmarks}
\label{app:coding}

The nine benchmarks in the main paper are single-turn reasoning tasks, where
an agent emits one answer per step. To test whether the step-wise routing of
\ourmethod still pays off when each step is a long-horizon, stateful
interaction with an execution environment, we further evaluate on six
agentic coding and command-line benchmarks.

\noindent\textbf{Benchmarks.} We select benchmarks that differ along two
axes, the length of the interaction horizon and whether the agent acts on a
shell or on a repository:
\begin{itemize}[leftmargin=1.2em,itemsep=0pt,topsep=2pt,parsep=0pt]
    \item \textbf{Terminal-Bench 2.0}: 89 end-to-end tasks solved through a
    real terminal, scored by task-specific verification scripts;
    \item \textbf{SWE-bench Verified}: 500 human-validated GitHub issues,
    scored by the fraction of instances whose test suite turns green;
    \item \textbf{SWE-Lancer} (IC SWE Diamond): freelance software tasks,
    scored by the payout of the passed tasks;
    \item \textbf{Aider Polyglot}: multi-language edit-and-repair tasks,
    scored by pass rate;
    \item \textbf{BigCodeBench-Hard} (Complete split): compositional library
    usage, scored by Pass@1;
    \item \textbf{LiveCodeBench v6}: contamination-controlled competitive
    programming, scored by Pass@1.
\end{itemize}

\noindent\textbf{Baselines.} Besides the general-purpose systems of
Table~\ref{tab: main results}, we add the coding-specific scaffolds that
dominate these leaderboards: ReAct, CodeAct, SWE-agent, and
OpenHands. All systems share the same backbone
and the same agent budget as in Appendix~\ref{app:baseline}; for \ourmethod
the agent pool is replaced with the task-specific coding pool of
Appendix~\ref{app:agent_pool}.

\begin{table*}[t]
    \centering
    \renewcommand\tabcolsep{5.3pt}
    \renewcommand\arraystretch{1.1}
    \resizebox{\linewidth}{!}{
    \begin{tabular}{lccccccc}
    \toprule
    \rowcb
        \textbf{Methods} & \textbf{Terminal-Bench 2.0} & \textbf{SWE-bench Verified} & \textbf{SWE-Lancer} & \textbf{Aider Polyglot} & \textbf{BigCodeBench-Hard} & \textbf{LiveCodeBench v6} & \textbf{Avg.} \\ \midrule
        Vanilla & 17.80 & 26.00 & 14.50 & 41.80 & 28.40 & 87.80 & 36.05 \\ \hline
        \multicolumn{8}{l}{\textit{\textbf{\small Single-Agent Coding Scaffolds}}} \\
        \rowcg
        ReAct & 20.60\red{2.80} & 34.34\red{8.34} & 18.00\red{3.50} & 44.29\red{2.49} & 32.18\red{3.78} & 88.55\red{0.75} & 39.66 \\
        CodeAct & 25.61\red{7.81} & 37.88\red{11.88} & 21.73\red{7.23} & 46.10\red{4.30} & 34.28\red{5.88} & 88.15\red{0.35} & 42.29 \\
        \rowcg
        SWE-agent & 29.18\red{11.38} & 47.35\red{21.35} & 23.48\red{8.98} & 45.77\red{3.97} & 35.81\red{7.41} & 89.11\red{1.31} & 45.12 \\
        OpenHands & 30.29\red{12.49} & 48.50\red{22.50} & 21.65\red{7.15} & 46.09\red{4.29} & 35.71\red{7.31} & 89.42\red{1.62} & 45.28 \\ \hline
        \multicolumn{8}{l}{\textit{\textbf{\small Multi-Agent Systems}}} \\
        \rowcg
        ChatDev & 20.27\red{2.47} & 33.77\red{7.77} & 18.95\red{4.45} & 44.95\red{3.15} & 31.10\red{2.70} & 88.56\red{0.76} & 39.60 \\
        MetaGPT & 24.61\red{6.81} & 37.94\red{11.94} & 19.79\red{5.29} & 45.12\red{3.32} & 31.19\red{2.79} & 89.13\red{1.33} & 41.30 \\
        \rowcg
        AutoGen & 25.39\red{7.59} & 38.11\red{12.11} & 23.84\red{9.34} & 42.58\red{0.78} & 34.31\red{5.91} & 88.78\red{0.98} & 42.17 \\
        LLM-Debate & 23.71\red{5.91} & 42.64\red{16.64} & 20.01\red{5.51} & 46.63\red{4.83} & 32.57\red{4.17} & 88.62\red{0.82} & 42.36 \\
        \rowcg
        MoA & 24.23\red{6.43} & 40.94\red{14.94} & 21.08\red{6.58} & 45.03\red{3.23} & 32.98\red{4.58} & 89.18\red{1.38} & 42.24 \\ \hline
        \multicolumn{8}{l}{\textit{\textbf{\small Self-Evolving Multi-Agent Systems}}} \\
        \rowcg
        GPTSwarm & 28.80\red{11.00} & 45.66\red{19.66} & 24.73\red{10.23} & 46.56\red{4.76} & 36.09\red{7.69} & 90.14\red{2.34} & 45.33 \\
        G-Designer & \second{34.06}\red{16.26} & 49.75\red{23.75} & 24.88\red{10.38} & 48.75\red{6.95} & 37.91\red{9.51} & 89.53\red{1.73} & 47.48 \\
        \rowcg
        ARG-Designer & 30.10\red{12.30} & 51.10\red{25.10} & \second{27.07}\red{12.57} & 48.72\red{6.92} & 36.34\red{7.94} & 89.57\red{1.77} & 47.15 \\
        SafeSieve & 33.78\red{15.98} & \second{52.36}\red{26.36} & 25.75\red{11.25} & \second{48.97}\red{7.17} & \second{38.84}\red{10.44} & 89.99\red{2.19} & \second{48.28} \\
        \rowcg
        AFlow & 29.24\red{11.44} & 47.08\red{21.08} & 23.48\red{8.98} & 46.77\red{4.97} & 37.27\red{8.87} & 89.76\red{1.96} & 45.60 \\
        STEER & 32.99\red{15.19} & 48.40\red{22.40} & 26.39\red{11.89} & 47.00\red{5.20} & 38.36\red{9.96} & \second{90.40}\red{2.60} & 47.26 \\ \hline
        \rowcr
        \ourmethod (Ours) & \first{40.03}\red{22.23} & \first{61.44}\red{35.44} & \first{29.70}\red{15.20} & \first{51.57}\red{9.77} & \first{43.04}\red{14.64} & \first{90.97}\red{3.17} & \first{52.79} \\
        \bottomrule
    \end{tabular}}
    \caption{ Comparison on agentic coding and command-line benchmarks, all driven by \llmname{gpt-oss-120b} (Vanilla). Terminal-Bench, SWE-bench Verified and Aider Polyglot report resolve rate (\%), BigCodeBench-Hard and LiveCodeBench report Pass@1 (\%), and SWE-Lancer reports the earned payout normalized to the full pool (\%). Subscripts are the gain over Vanilla. \first{Red}: the best, \second{Blue}: the runner-up.}
    \label{tab:coding}
\end{table*}

\noindent\textbf{Results.} \ourmethod leads on all six benchmarks, averaging
$52.79$ against $36.05$ for Vanilla and $48.28$ for SafeSieve, the strongest
baseline on average. The margin over the best baseline of each benchmark
is not uniform, and it tracks the
length of the interaction horizon: it is widest on SWE-bench Verified
($+9.08$) and Terminal-Bench 2.0 ($+5.97$), the two suites where a single
episode spans dozens of tool calls, and narrows to $+2.60$ on Aider
Polyglot and $+2.63$ on SWE-Lancer, where an episode is closer to a single
scoped edit. The two non-interactive suites sit outside this trend, at
$+4.20$ on BigCodeBench-Hard and $+0.57$ on the saturated LiveCodeBench v6.
This is the behavior the step-wise router is designed for:
these benchmarks reward an agent that can change its own composition
mid-trajectory, e.g.\ switching from a planner to a debugger once a test
fails, whereas a pre-compiled topology must commit to that decision before
the first command is run.

\noindent\textbf{Where the baselines stand.} The ordering of the baselines
departs from Table~\ref{tab: main results} in one systematic way. The
coding-specific scaffolds are far more competitive here than general
multi-agent systems of comparable cost: SWE-agent and OpenHands average
$45.12$ and $45.28$, above every general multi-agent system, and the gap
is widest on SWE-bench Verified, where they reach $47.35$ and $48.50$
against $42.64$ for the strongest general system. Repository navigation
priors matter on that column in a way they do not on the reasoning
benchmarks of the main paper. \ourmethod leads it at $61.44$, which suggests the two mechanisms are complementary rather
than redundant.

\noindent\textbf{How much is left.} Averaged over the six benchmarks,
\ourmethod closes $26.0\%$ of the gap between Vanilla and a perfect score,
ranging from $16.8\%$ to $47.9\%$. The comparable figure on the nine
reasoning benchmarks is $64.7\%$, so agentic coding remains far
from saturated and leaves considerably more room than the main results
suggest. Two columns should be read with care. LiveCodeBench v6 starts at
$87.80$ for Vanilla and moves only $+3.17$, so it separates systems poorly
and mainly serves as a non-regression check. SWE-Lancer sits at the other
extreme, with absolute numbers low enough that run-to-run variance may
dominate, so we report the mean over the five runs of
Appendix~\ref{app:reproducibility}.

% ---------------------------------------------------------------------
\subsection{Generalization Across Base Models}
\label{app:backbones}

The main results are obtained with \llmname{gpt-oss-120b}. To verify that the
gains come from the routing mechanism rather than from one particular
backbone, we repeat the full comparison of Table~\ref{tab: main results} with
two further backbones, \llmname{qwen3.8-27b} (a 27B dense model) and
\llmname{deepseek-v4-flash} (a sparse model with 13B activated parameters).
All agents in the pool, all baselines, and \ourmethod use the same backbone
within a table; every other hyperparameter follows Appendix~\ref{app:hyper}.

\begin{table*}[t]
    \centering
    \renewcommand\tabcolsep{5.3pt}
    \renewcommand\arraystretch{1.1}
    \resizebox{\linewidth}{!}{
    \begin{tabular}{lcccccccccc}
    \toprule
    \rowcb
        \textbf{Methods} & \textbf{MMLU} & \textbf{GSM8K} & \textbf{MultiArith} & \textbf{SVAMP} & \textbf{AQuA} & \textbf{HumanEval} & \textbf{DS-1000} & \textbf{HotpotQA} & \textbf{DDXPlus} & \textbf{Avg.} \\ \midrule
        Vanilla & 78.15 & 88.60 & 94.25 & 88.40 & 72.85 & 79.45 & 42.70 & 79.20 & 54.85 & 75.38 \\ \hline
        \multicolumn{11}{l}{\textit{\textbf{\small Single-Agent Systems}}} \\
        \rowcg
        CoT & 80.59\red{2.44} & 89.67\red{1.07} & 94.61\red{0.36} & 89.58\red{1.18} & 78.83\red{5.98} & 80.00\red{0.55} & 51.06\red{8.36} & 77.52\blue{1.68} & 63.33\red{8.48} & 78.35 \\
        SC & 83.18\red{5.03} & 88.78\red{0.18} & 95.78\red{1.53} & 91.03\red{2.63} & 79.37\red{6.52} & 82.68\red{3.23} & 44.57\red{1.87} & 81.86\red{2.66} & 63.60\red{8.75} & 78.98 \\ \hline
        \multicolumn{11}{l}{\textit{\textbf{\small Multi-Agent Systems}}} \\
        \rowcg
        Chain & 79.27\red{1.12} & 88.04\blue{0.56} & 93.56\blue{0.69} & 89.04\red{0.64} & 75.21\red{2.36} & 80.42\red{0.97} & 45.99\red{3.29} & 76.57\blue{2.63} & 61.21\red{6.36} & 76.59 \\
        Tree & 79.06\red{0.91} & 87.17\blue{1.43} & 94.08\blue{0.17} & 88.73\red{0.33} & 70.59\blue{2.26} & 77.51\blue{1.94} & 48.32\red{5.62} & 77.07\blue{2.13} & 70.70\red{15.85} & 77.03 \\
        \rowcg
        Star & 76.74\blue{1.41} & 85.69\blue{2.91} & 94.39\red{0.14} & 89.54\red{1.14} & 71.86\blue{0.99} & 80.32\red{0.87} & 46.47\red{3.77} & 79.76\red{0.56} & 63.79\red{8.94} & 76.51 \\
        Complete Graph & 78.81\red{0.66} & 87.95\blue{0.65} & 95.14\red{0.89} & 87.15\blue{1.25} & 70.00\blue{2.85} & 86.46\red{7.01} & 47.44\red{4.74} & 78.70\blue{0.50} & 68.02\red{13.17} & 77.74 \\
        \rowcg
        Random Graph & 81.06\red{2.91} & 88.77\red{0.17} & 94.89\red{0.64} & 87.58\blue{0.82} & 70.61\blue{2.24} & 89.58\red{10.13} & 54.67\red{11.97} & 80.31\red{1.11} & 70.03\red{15.18} & 79.72 \\
        AutoGen & 80.19\red{2.04} & 90.38\red{1.78} & 95.03\red{0.78} & 86.78\blue{1.62} & 72.62\blue{0.23} & 87.68\red{8.23} & 52.77\red{10.07} & 79.33\red{0.13} & 68.06\red{13.21} & 79.20 \\
        \rowcg
        LLM-Debate & 81.63\red{3.48} & 91.34\red{2.74} & 96.16\red{1.91} & 89.62\red{1.22} & 80.26\red{7.41} & 87.74\red{8.29} & 51.77\red{9.07} & 82.41\red{3.21} & 71.34\red{16.49} & 81.36 \\
        MoA & 80.45\red{2.30} & 91.13\red{2.53} & 95.96\red{1.71} & 90.22\red{1.82} & 74.77\red{1.92} & 87.52\red{8.07} & 41.47\blue{1.23} & 81.34\red{2.14} & 72.98\red{18.13} & 79.54 \\ \hline
        \multicolumn{11}{l}{\textit{\textbf{\small Self-Evolving Multi-Agent Systems}}} \\
        \rowcg
        GPTSwarm & 82.73\red{4.58} & 92.03\red{3.43} & 97.62\red{3.37} & 91.96\red{3.56} & 81.21\red{8.36} & 90.67\red{11.22} & 57.76\red{15.06} & 83.00\red{3.80} & 75.52\red{20.67} & 83.61 \\
        G-Designer & 82.29\red{4.14} & 94.60\red{6.00} & 97.96\red{3.71} & 91.49\red{3.09} & 77.89\red{5.04} & 90.89\red{11.44} & 58.38\red{15.68} & 85.00\red{5.80} & 74.57\red{19.72} & 83.67 \\
        \rowcg
        ARG-Designer & 83.09\red{4.94} & 94.16\red{5.56} & 97.26\red{3.01} & 88.86\red{0.46} & 77.28\red{4.43} & 87.05\red{7.60} & 57.49\red{14.79} & 82.08\red{2.88} & 69.21\red{14.36} & 81.83 \\
        SafeSieve & 82.94\red{4.79} & \second{95.25}\red{6.65} & \second{98.11}\red{3.86} & \second{92.10}\red{3.70} & 82.93\red{10.08} & 91.49\red{12.04} & \second{58.76}\red{16.06} & 86.00\red{6.80} & 77.98\red{23.13} & \second{85.06} \\
        \rowcg
        AFlow & 78.55\red{0.40} & 90.74\red{2.14} & 95.08\red{0.83} & 86.17\blue{2.23} & 80.82\red{7.97} & 89.75\red{10.30} & 49.96\red{7.26} & 86.11\red{6.91} & 73.10\red{18.25} & 81.14 \\
        SpecReason & 83.60\red{5.45} & 92.95\red{4.35} & 96.35\red{2.10} & 89.91\red{1.51} & 83.34\red{10.49} & \second{91.65}\red{12.20} & 53.48\red{10.78} & \second{87.89}\red{8.69} & \second{79.55}\red{24.70} & 84.30 \\
        \rowcg
        STEER & \second{83.63}\red{5.48} & 93.61\red{5.01} & 96.25\red{2.00} & 88.77\red{0.37} & \second{86.04}\red{13.19} & 90.91\red{11.46} & 53.60\red{10.90} & 85.96\red{6.76} & 72.70\red{17.85} & 83.50 \\ \hline
        \rowcr
        \ourmethod (Ours) & \first{90.67}\red{12.52} & \first{98.90}\red{10.30} & \first{99.27}\red{5.02} & \first{95.83}\red{7.43} & \first{88.25}\red{15.40} & \first{95.78}\red{16.33} & \first{65.39}\red{22.69} & \first{89.97}\red{10.77} & \first{87.52}\red{32.67} & \first{90.18} \\
        \bottomrule
    \end{tabular}}
    \caption{ Accuracy on the nine main benchmarks when all systems are driven by \llmname{qwen3.8-27b}. Subscripts are the gain over Vanilla. \first{Red}: the best, \second{Blue}: the runner-up.}
    \label{tab:backbone_qwen}
\end{table*}

\begin{table*}[t]
    \centering
    \renewcommand\tabcolsep{5.3pt}
    \renewcommand\arraystretch{1.1}
    \resizebox{\linewidth}{!}{
    \begin{tabular}{lcccccccccc}
    \toprule
    \rowcb
        \textbf{Methods} & \textbf{MMLU} & \textbf{GSM8K} & \textbf{MultiArith} & \textbf{SVAMP} & \textbf{AQuA} & \textbf{HumanEval} & \textbf{DS-1000} & \textbf{HotpotQA} & \textbf{DDXPlus} & \textbf{Avg.} \\ \midrule
        Vanilla & 82.90 & 89.85 & 94.80 & 89.60 & 74.40 & 82.15 & 46.25 & 84.10 & 58.75 & 78.09 \\ \hline
        \multicolumn{11}{l}{\textit{\textbf{\small Single-Agent Systems}}} \\
        \rowcg
        CoT & 82.40\blue{0.50} & 90.28\red{0.43} & 94.62\blue{0.18} & 89.12\blue{0.48} & 75.09\red{0.69} & 80.51\blue{1.64} & 50.95\red{4.70} & 81.18\blue{2.92} & 65.87\red{7.12} & 78.89 \\
        SC & 85.93\red{3.03} & 89.98\red{0.13} & 95.47\red{0.67} & 89.62\red{0.02} & 74.62\red{0.22} & 79.01\blue{3.14} & 43.91\blue{2.34} & 86.28\red{2.18} & 66.93\red{8.18} & 79.08 \\ \hline
        \multicolumn{11}{l}{\textit{\textbf{\small Multi-Agent Systems}}} \\
        \rowcg
        Chain & 85.26\red{2.36} & 89.07\blue{0.78} & 94.12\blue{0.68} & 87.91\blue{1.69} & 70.68\blue{3.72} & 83.58\red{1.43} & 48.43\red{2.18} & 83.51\blue{0.59} & 68.48\red{9.73} & 79.00 \\
        Tree & 83.86\red{0.96} & 89.28\blue{0.57} & 95.04\red{0.24} & 91.38\red{1.78} & 75.67\red{1.27} & 87.29\red{5.14} & 51.22\red{4.97} & 81.92\blue{2.18} & 73.33\red{14.58} & 81.00 \\
        \rowcg
        Star & 82.64\blue{0.26} & 87.08\blue{2.77} & 94.62\blue{0.18} & 89.38\blue{0.22} & 73.64\blue{0.76} & 81.31\blue{0.84} & 51.04\red{4.79} & 84.70\red{0.60} & 68.39\red{9.64} & 79.20 \\
        Complete Graph & 84.00\red{1.10} & 90.50\red{0.65} & 95.49\red{0.69} & 89.78\red{0.18} & 75.09\red{0.69} & 88.26\red{6.11} & 51.81\red{5.56} & 85.06\red{0.96} & 68.09\red{9.34} & 80.90 \\
        \rowcg
        Random Graph & 87.26\red{4.36} & 92.14\red{2.29} & 96.00\red{1.20} & 89.53\blue{0.07} & 76.49\red{2.09} & 88.53\red{6.38} & 55.87\red{9.62} & 87.55\red{3.45} & 71.80\red{13.05} & 82.80 \\
        AutoGen & \second{88.07}\red{5.17} & 92.58\red{2.73} & 96.31\red{1.51} & 91.37\red{1.77} & 74.69\red{0.29} & 90.12\red{7.97} & 53.28\red{7.03} & 87.42\red{3.32} & 74.31\red{15.56} & 83.13 \\
        \rowcg
        LLM-Debate & 85.17\red{2.27} & 89.66\blue{0.19} & 96.03\red{1.23} & 89.58\blue{0.02} & 72.60\blue{1.80} & 90.20\red{8.05} & 55.35\red{9.10} & 84.58\red{0.48} & 72.84\red{14.09} & 81.78 \\
        MoA & 87.01\red{4.11} & 90.90\red{1.05} & 95.37\red{0.57} & 90.21\red{0.61} & 70.87\blue{3.53} & 90.24\red{8.09} & 48.06\red{1.81} & 85.62\red{1.52} & 69.96\red{11.21} & 80.92 \\ \hline
        \multicolumn{11}{l}{\textit{\textbf{\small Self-Evolving Multi-Agent Systems}}} \\
        \rowcg
        GPTSwarm & 85.65\red{2.75} & 91.55\red{1.70} & 96.47\red{1.67} & 88.01\blue{1.59} & 75.12\red{0.72} & 93.77\red{11.62} & \second{61.12}\red{14.87} & 87.84\red{3.74} & 77.50\red{18.75} & 84.11 \\
        G-Designer & 85.43\red{2.53} & 94.05\red{4.20} & 97.42\red{2.62} & 89.97\red{0.37} & 78.82\red{4.42} & 89.41\red{7.26} & 55.53\red{9.28} & 87.98\red{3.88} & 78.71\red{19.96} & 84.15 \\
        \rowcg
        ARG-Designer & 84.59\red{1.69} & \second{96.09}\red{6.24} & \second{98.26}\red{3.46} & 92.33\red{2.73} & \second{86.58}\red{12.18} & 93.40\red{11.25} & 58.86\red{12.61} & 85.66\red{1.56} & \second{79.97}\red{21.22} & 86.19 \\
        SafeSieve & 87.64\red{4.74} & 94.48\red{4.63} & 98.12\red{3.32} & \second{92.58}\red{2.98} & 83.65\red{9.25} & 92.04\red{9.89} & 58.27\red{12.02} & \second{90.68}\red{6.58} & 79.66\red{20.91} & \second{86.35} \\
        \rowcg
        AFlow & 84.36\red{1.46} & 90.05\red{0.20} & 94.53\blue{0.27} & 87.28\blue{2.32} & 79.27\red{4.87} & 90.15\red{8.00} & 48.31\red{2.06} & 90.06\red{5.96} & 74.44\red{15.69} & 82.05 \\
        SpecReason & 87.24\red{4.34} & 92.48\red{2.63} & 96.35\red{1.55} & 90.03\red{0.43} & 86.41\red{12.01} & \second{94.86}\red{12.71} & 52.94\red{6.69} & 90.08\red{5.98} & 76.46\red{17.71} & 85.21 \\
        \rowcg
        STEER & 84.33\red{1.43} & 94.51\red{4.66} & 96.71\red{1.91} & 91.06\red{1.46} & 84.10\red{9.70} & 90.97\red{8.82} & 53.09\red{6.84} & 89.90\red{5.80} & 73.14\red{14.39} & 84.20 \\ \hline
        \rowcr
        \ourmethod (Ours) & \first{93.15}\red{10.25} & \first{98.64}\red{8.79} & \first{99.11}\red{4.31} & \first{95.96}\red{6.36} & \first{88.13}\red{13.73} & \first{96.60}\red{14.45} & \first{67.57}\red{21.32} & \first{92.96}\red{8.86} & \first{83.86}\red{25.11} & \first{90.66} \\
        \bottomrule
    \end{tabular}}
    \caption{ Accuracy on the nine main benchmarks when all systems are driven by \llmname{deepseek-v4-flash}. Subscripts are the gain over Vanilla. \first{Red}: the best, \second{Blue}: the runner-up.}
    \label{tab:backbone_deepseek}
\end{table*}

To make the cross-backbone trend readable at a glance, we also condense the
three tables into the average accuracy of each system under each backbone.

\begin{table}[t]
    \centering
    \small
    \renewcommand\arraystretch{1.1}
    \resizebox{\columnwidth}{!}{
    \begin{tabular}{lccc}
    \toprule
    \rowcb
        \textbf{Methods} & \llmname{gpt-oss-120b} & \llmname{qwen3.8-27b} & \llmname{deepseek-v4-flash} \\
        \midrule
        Vanilla & 74.22 & 75.38 & 78.09 \\
        \rowcg
        LLM-Debate & 80.10 & 81.36 & 81.78 \\
        MoA & 77.71 & 79.54 & 80.92 \\
        \rowcg
        G-Designer & 83.30 & 83.67 & 84.15 \\
        ARG-Designer & 83.51 & 81.83 & 86.19 \\
        \rowcg
        SafeSieve & \second{83.91} & \second{85.06} & \second{86.35} \\
        STEER & 83.19 & 83.50 & 84.20 \\ \hline
        \rowcr
        \ourmethod (Ours) & \first{89.38} & \first{90.18} & \first{90.66} \\ \hline
        \textit{Margin over runner-up} & \textit{+5.47} & \textit{+5.12} & \textit{+4.31} \\
        \bottomrule
    \end{tabular}}
    \caption{ Average accuracy across the nine main benchmarks under three backbones. The \llmname{gpt-oss-120b} column is copied from Table~\ref{tab: main results}, and the other two are condensed from Tables~\ref{tab:backbone_qwen} and~\ref{tab:backbone_deepseek}.}
    \label{tab:backbone_summary}
\end{table}

\noindent\textbf{Results.} \ourmethod is the best system on all nine columns
under both backbones, reaching $90.18$ with \llmname{qwen3.8-27b} and
$90.66$ with \llmname{deepseek-v4-flash}, so the advantage is a property of
the routing mechanism rather than of one particular model. The margin over
the strongest baseline narrows monotonically as the backbone strengthens,
from $+5.47$ with \llmname{gpt-oss-120b} to $+5.12$ and then $+4.31$, while
the Vanilla level rises from $74.22$ to $75.38$ and $78.09$. Adaptive
routing therefore behaves more like a compensator for a weaker agent than
like an amplifier of a strong one.

\noindent\textbf{Where the gain concentrates.} Within each backbone the gain
is strongly anti-correlated with how well the single agent already does: the
Pearson correlation between the Vanilla score of a benchmark and the gain of
\ourmethod on it is $-0.87$ under \llmname{qwen3.8-27b} and $-0.92$ under
\llmname{deepseek-v4-flash}. Concretely, the largest gains are on DDXPlus
($+32.67$ and $+25.11$) and DS-1000 ($+22.69$ and $+21.32$), the two
benchmarks where a single agent is weakest, and the smallest is on
MultiArith ($+5.02$ and $+4.31$), which already starts above $94$. Expressed
as a fraction of the distance between Vanilla and a perfect score,
\ourmethod closes $66.5\%$ on average under \llmname{qwen3.8-27b} and
$64.6\%$ under \llmname{deepseek-v4-flash}, with per-benchmark
ranges of $39.6\%$--$90.4\%$ and $39.7\%$--$86.6\%$.

\noindent\textbf{Stability of the baselines.} SafeSieve is the strongest
baseline on average under all three backbones, but it does not hold that
position column by column: STEER leads MMLU and AQuA under
\llmname{qwen3.8-27b}, SpecReason leads HumanEval, HotpotQA and DDXPlus, and
under \llmname{deepseek-v4-flash} ARG-Designer takes four of the nine
columns while AutoGen takes MMLU. Several baselines sit within one point of
each other, so single-run comparisons among them are not meaningful and
should be read together with the variance discussed in
Appendix~\ref{app:reproducibility}.

% ---------------------------------------------------------------------
\subsection{Ablation on the Semantic Encoder}
\label{app:encoder}

\ourmethod compresses the query and every intermediate response into a
fixed-size vector through $\texttt{ST}(\cdot)$, which by default is
\texttt{all-MiniLM-L6-v2} with $D = 384$. Since this encoder is the only
component that decides what the router actually sees, we ablate it along
three dimensions: encoder capacity, embedding dimension $D$, and whether
semantic structure is needed at all.

\noindent\textbf{Variants.} We cover a $22$M-to-$1.5$B capacity range, and
add two controls. \texttt{TF-IDF + random projection} keeps the dimension
but destroys contextual semantics, and \texttt{GloVe mean pooling} keeps
lexical semantics but discards word order. We further reuse the last-layer
hidden state of the agent LLM itself, which is the direction suggested in
Appendix~\ref{app:limitations}, both frozen and with a trainable linear head
that projects it back to $384$ dimensions. The router and all other
hyperparameters are held fixed; only $\texttt{ST}$ and the input width of
the RNN-Router change.

\begin{table*}[t]
    \centering
    \renewcommand\tabcolsep{5.3pt}
    \renewcommand\arraystretch{1.1}
    \resizebox{\linewidth}{!}{
    \begin{tabular}{lcccccccc}
    \toprule
    \rowcb
        \textbf{Semantic Encoder} & \textbf{Params} & \textbf{$D$} & \textbf{MMLU} & \textbf{GSM8K} & \textbf{HumanEval} & \textbf{DS-1000} & \textbf{Avg.} & \textbf{Latency (ms)} \\ \midrule
        \multicolumn{9}{l}{\textit{\textbf{\small Controls}}} \\
        \rowcg
        TF-IDF + random projection & -- & 384 & 85.71 & 96.31 & 87.00 & 47.00 & 79.00 & 0.4 \\
        GloVe mean pooling & -- & 300 & 87.42 & 96.75 & 90.44 & 51.77 & 81.59 & 0.3 \\ \hline
        \multicolumn{9}{l}{\textit{\textbf{\small Sentence Encoders}}} \\
        \rowcr
        \texttt{all-MiniLM-L6-v2} (default) & 22M & 384 & \first{91.35} & \first{98.87} & \first{95.62} & \first{64.34} & \first{87.55} & 3.1 \\
        \rowcg
        \texttt{all-mpnet-base-v2} & 110M & 768 & 90.68 & 98.76 & \second{95.37} & 63.51 & 87.08 & 8.7 \\
        \texttt{e5-large-v2} & 335M & 1024 & 90.62 & 98.52 & 94.36 & 61.96 & 86.36 & 18.3 \\
        \rowcg
        \texttt{bge-m3} & 568M & 1024 & 91.10 & 98.68 & 95.32 & \second{63.80} & \second{87.23} & 26.5 \\
        \texttt{Qwen3-Embedding-0.6B} & 0.6B & 1024 & 91.02 & 98.74 & 94.85 & 63.32 & 86.98 & 31.2 \\
        \rowcg
        \texttt{gte-Qwen2-1.5B-instruct} & 1.5B & 1536 & \second{91.24} & 98.62 & 94.28 & 63.43 & 86.89 & 74.6 \\ \hline
        \multicolumn{9}{l}{\textit{\textbf{\small Reusing the Agent Backbone}}} \\
        \rowcg
        Backbone hidden state (frozen) & -- & 2880 & 89.73 & 98.10 & 92.83 & 57.35 & 84.50 & 142.0 \\
        Backbone hidden state + linear head & +1.1M & 384 & 91.09 & \second{98.77} & 94.62 & 63.64 & 87.03 & 148.3 \\
        \bottomrule
    \end{tabular}}
    \caption{ Ablation on the semantic encoder $\texttt{ST}(\cdot)$ of \ourmethod, with \llmname{gpt-oss-120b} as the backbone. The default row is copied from Table~\ref{tab: main results}. Params counts the encoder only, and the agent LLMs are unchanged. Latency is the mean wall-clock time of one encoding call on a single A100. \first{Red}: the best, \second{Blue}: the runner-up.}
    \label{tab:encoder_ablation}
\end{table*}

\noindent\textbf{Results.} The default configuration is the best on every
column and on the average, at $87.55$ against $87.23$ for the runner-up.
Both controls fall far behind, losing $8.55$ and $5.96$ average points, so
the router is reading semantic content rather than merely exploiting step
position. The informative part is that no larger encoder helps either:
\texttt{gte-Qwen2-1.5B-instruct} has $68\times$ the parameters of the
default and still lands $0.66$ points below it, and \texttt{e5-large-v2} is
the weakest sentence encoder of the group at $86.36$ despite being $15\times$
larger than the default.

\noindent\textbf{Why capacity does not help.} The pattern follows $D$ rather
than parameter count, which follows because $D$ is also the input width
of the RNN-Router: a wider encoder enlarges the only module we train, and
the routing signal is too sparse to fit it. The clearest evidence is the
last block. Taking the frozen backbone hidden state at $D = 2880$ is the
worst learned variant at $84.50$, but attaching a linear head that projects
it back to $384$ recovers $2.53$ of that $3.05$-point gap, reaching $87.03$
without changing the underlying representation at all. The dimension, not
the quality of the embedding, is what the router is sensitive to. Since the
default is also the cheapest learned encoder by a wide margin, at $3.1$\,ms per
encoding call against $74.6$\,ms for the $1.5$B encoder and $148.3$\,ms for
the backbone head, there is no configuration in this table that trades
favorably against it.

\section{Reproducibility and Statistical Considerations}
\label{app:reproducibility}

\paragraph{Evaluation stability.}
Our main experiments are designed to provide a stable measurement of model performance under mildly stochastic decoding.
Specifically, all methods are instantiated with the same backbone model, \texttt{gpt-oss-120b}, and we set the decoding temperature to $0.1$ for all evaluations.
For each benchmark, we independently run the evaluation five times under the same prompts, router parameters, benchmark splits, and evaluation scripts, and report the mean accuracy together with the standard deviation in Table~\ref{tab:stability}.

This protocol allows us to quantify the small variance induced by low-temperature decoding while keeping the evaluation setting highly controlled.
The reported results therefore reflect both the average performance and the run-to-run stability of \ourmethod.
Since the decoding temperature is close to deterministic, the observed variance remains small, indicating that the main performance gain comes from the routing mechanism and adaptive multi-agent collaboration, rather than from sampling diversity.

We nevertheless acknowledge that exact reproducibility in LLM systems can still be affected by external factors such as backend model updates, API-side implementation differences, or tool-side nondeterminism.
To minimize such effects, we keep the backbone fixed, use the same decoding temperature throughout evaluation, and standardize the tool interface and prompting format across all baselines and \ourmethod.

\begin{table}[!htbp]
\centering
\small
\resizebox{\linewidth}{!}{
\begin{tabular}{lcccccccccc}
\toprule
\textbf{Method} & \textbf{MMLU} & \textbf{GSM8K} & \textbf{MultiArith} & \textbf{SVAMP} & \textbf{AQuA} & \textbf{HumanEval} & \textbf{DS-1000} & \textbf{HotpotQA} & \textbf{DDXPlus} & \textbf{Avg.} \\
\midrule
\ourmethod 
& $91.35{\scriptstyle \pm 0.19}$ 
& $98.87{\scriptstyle \pm 0.13}$ 
& $99.15{\scriptstyle \pm 0.09}$ 
& $94.76{\scriptstyle \pm 0.22}$ 
& $86.60{\scriptstyle \pm 0.34}$ 
& $95.62{\scriptstyle \pm 0.28}$ 
& $64.34{\scriptstyle \pm 0.46}$ 
& $90.38{\scriptstyle \pm 0.31}$ 
& $83.37{\scriptstyle \pm 0.41}$ 
& $89.38{\scriptstyle \pm 0.24}$ \\
\bottomrule
\end{tabular}}
\caption{Evaluation stability of \ourmethod over five independent runs. 
All experiments use \texttt{gpt-oss-120b} as the backbone model with temperature $0.1$. 
We report mean accuracy and standard deviation.}
\label{tab:stability}
\end{table}

\subsection{Broader Impacts}
\label{app:impact}

This work has the potential to bring positive impact to a range of real-world applications.
By enabling more adaptive and efficient multi-agent reasoning, \ourmethod may benefit domains such as education, scientific assistance, coding support, and general decision-support systems.
More broadly, it contributes to the study of collaborative intelligence in LLM-based systems and may facilitate the development of more effective and resource-efficient agentic frameworks.

At the same time, as with other general-purpose LLM-based systems, careful evaluation remains important when deploying such methods in real-world scenarios, especially in domains requiring high reliability.
We therefore recommend benchmark-based validation, transparent reporting of system behavior, and appropriate human oversight in safety-sensitive applications.

\subsection{The Use of Large Language Models}
\label{app:llm_usage}

LLMs are used as an essential methodological component of this work.
Our proposed \ourmethod is a multi-agent framework in which agents are instantiated by LLM backbones, role profiles, and tool interfaces, and all main experiments are conducted based on this setup.
In our experiments, \llmname{gpt-oss-120b} is used as the unified backbone LLM for both \ourmethod and the compared baselines.

In addition, LLMs were used only for minor writing assistance, such as improving grammar and wording in the manuscript.
This auxiliary usage did not affect the method design, experimental results, or scientific conclusions of the paper.

\section{Limitations and Discussion}
\label{app:limitations}

Although \ourmethod achieves strong empirical performance, it still has several limitations.

\paragraph{Scalability of the agent pool.}
\ourmethod benefits from a diverse agent pool, but scaling the pool to a much larger number of agents may introduce additional routing difficulty.
As the candidate space grows, the router must discriminate among increasingly fine-grained role and tool combinations, which may require more adaptation data, stronger regularization, or more structured routing priors.
Moreover, a larger agent pool may also increase the optimization cost during training or test-time adaptation, since \ourmethod collects predictive entropy signals from candidate agents to supervise routing.

Therefore, exhaustively enumerating all combinations of LLM × Profile × Tools is inherently not an optimal solution. In the future work, we will explore novel mechanisms to avoid exhaustive enumeration while preserving the diversity of Agents.

\paragraph{Context-length sensitivity.}
Our routing mechanism relies on a Sentence Transformer to compress the query and intermediate responses into fixed-dimensional semantic representations.
While this design is lightweight and efficient, it may be less effective when the accumulated context becomes extremely long, highly heterogeneous, or densely interleaved across multiple reasoning steps.
In such cases, a single fixed-length semantic vector may not fully preserve the fine-grained structure of long-context dependencies, which can in turn affect routing quality. 

In the future, we will explore the potential of using the backbone of large LLMs to replace the Sentence Transformer, and training it like a reward model to make routing decisions.

\section{Showcases}

\begin{figure}[!htbp]
    \centering
\includegraphics[width=0.85\linewidth]{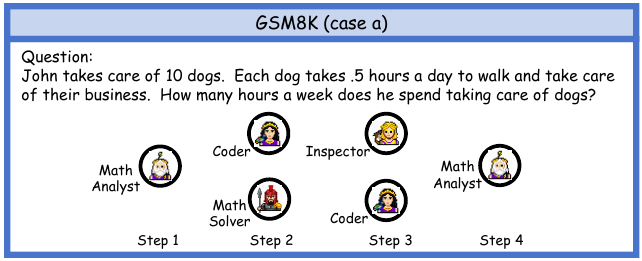}
    \caption{Case study on GSM8K.}
    \label{fig: case a}
\end{figure}

\begin{figure}[!htbp]
    \centering
\includegraphics[width=0.85\linewidth]{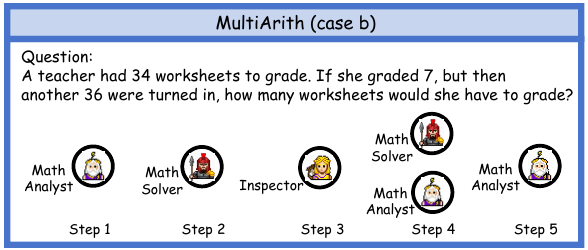}
    \caption{Case study on MultiArith.}
    \label{fig: case b}
    
\end{figure}

\begin{figure}[!htbp]
    \centering
\includegraphics[width=0.85\linewidth]{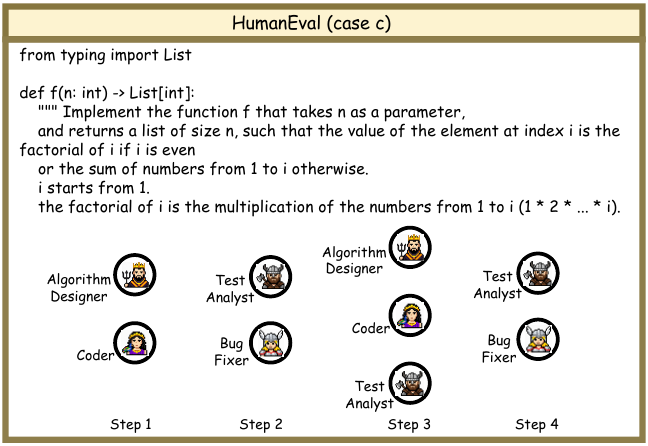}
    \caption{Case study on HumanEval.}
    \label{fig: case c}
    
\end{figure}

\begin{figure}[!htbp]
    \centering
\includegraphics[width=0.85\linewidth]{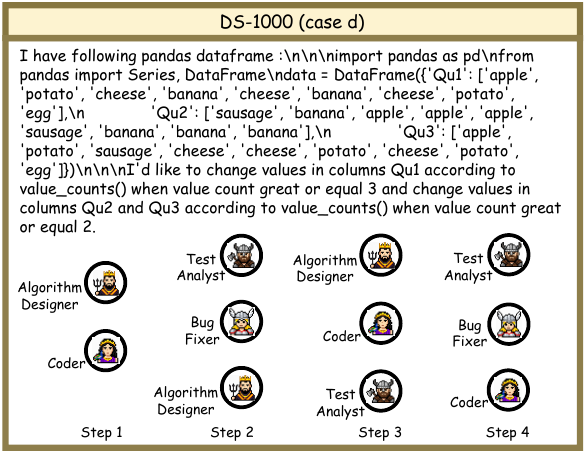}
    \caption{Case study on DS-1000.}
    \label{fig: case d}
    
\end{figure}

\begin{figure}[!htbp]
    \centering
\includegraphics[width=0.85\linewidth]{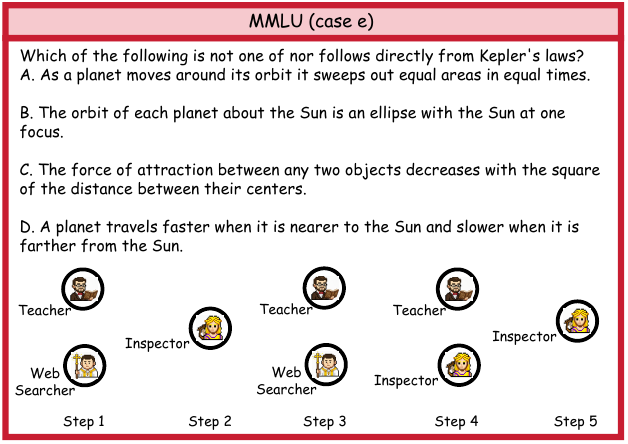}
    \caption{Case study on MMLU.}
    \label{fig: case e}
    
\end{figure}

We showcase some scheduling trajectories in this section--see Figure~\ref{fig: case a}--\ref{fig: case e}.

\newpage
\section{Details of Baselines}
For fair comparison, all baselines use \texttt{gpt-oss-120b}, the same prompt template as \ourmethod, and decoding temperature \(0.1\). 
For method-specific hyperparameters, we follow the original papers whenever available and tune validation-dependent thresholds under the same training budget of \(40\)--\(80\) queries.

\begin{table}[!htbp]
\centering
\small
\resizebox{0.65\columnwidth}{!}{
\begin{tabular}{lc}
\toprule
\textbf{Hyperparameter} & \textbf{Value} \\
\midrule
Backbone LLM & \texttt{gpt-oss-120b} \\
Prompt template & Same as \ourmethod \\
Decoding temperature & \(0.1\) \\
Optimization target & Graph edges and prompts \\
Graph optimizer & REINFORCE \\
Initial edge probability & \(0.10\) \\
Learning rate & \(0.40\) \\
Sampled graphs per iteration & \(20\) \\
Training budget & \(40\)--\(80\) queries \\
Stopping criterion & Budget exhausted \\
\bottomrule
\end{tabular}}
\caption{Key hyperparameter settings of GPTSwarm.}
\label{tab:gptswarm_hparams}
\end{table}

\begin{table}[!htbp]
\centering
\small
\resizebox{0.65\columnwidth}{!}{
\begin{tabular}{lc}
\toprule
\textbf{Hyperparameter} & \textbf{Value} \\
\midrule
Backbone LLM & \texttt{gpt-oss-120b} \\
Prompt template & Same as \ourmethod \\
Decoding temperature & \(0.1\) \\
Communication rounds & \(3\) \\
Node encoder & \texttt{all-MiniLM-L6-v2} \\
Node embedding dimension & \(384\) \\
Topology sampling times & \(10\) \\
Anchor regularization weight & \(0.01\) \\
Sparsity regularization weight & \(0.10\) \\
Training budget & \(40\)--\(80\) queries \\
\bottomrule
\end{tabular}}
\caption{Key hyperparameter settings of G-Designer.}
\label{tab:gdesigner_hparams}
\end{table}

\begin{table}[!htbp]
\centering
\small
\resizebox{0.65\columnwidth}{!}{
\begin{tabular}{lc}
\toprule
\textbf{Hyperparameter} & \textbf{Value} \\
\midrule
Backbone LLM & \texttt{gpt-oss-120b} \\
Prompt template & Same as \ourmethod \\
Decoding temperature & \(0.1\) \\
Search algorithm & Monte Carlo Tree Search \\
Workflow representation & Code-represented workflow \\
Optimized components & Workflow edges and node prompts \\
Workflow operator set & Original AFlow operators \\
Maximum optimization rounds & \(20\) \\
Training budget & \(40\)--\(80\) queries \\
Early stopping criterion & Validation score not improved \\
\bottomrule
\end{tabular}}
\caption{Key hyperparameter settings of AFlow.}
\label{tab:aflow_hparams}
\end{table}

\begin{table}[!htbp]
\centering
\small
\resizebox{0.65\columnwidth}{!}{
\begin{tabular}{lc}
\toprule
\textbf{Hyperparameter} & \textbf{Value} \\
\midrule
Backbone LLM & \texttt{gpt-oss-120b} \\
Prompt template & Same as \ourmethod \\
Decoding temperature & \(0.1\) \\
Routing granularity & Reasoning step \\
Draft model & \texttt{gpt-oss-120b} \\
Verifier model & \texttt{gpt-oss-120b} \\
Verification strategy & LLM-as-judge \\
Utility score range & \(0\) to \(9\) \\
Acceptance threshold candidates & \(3, 5, 7, 9\) \\
Acceptance threshold selection & Validation-selected \\
Fallback strategy & Regenerate rejected step \\
Training budget & \(40\)--\(80\) queries \\
\bottomrule
\end{tabular}}
\caption{Key hyperparameter settings of SpecReason.}
\label{tab:specreason_hparams}
\end{table}

\begin{table}[!htbp]
\centering
\small
\resizebox{0.65\columnwidth}{!}{
\begin{tabular}{lc}
\toprule
\textbf{Hyperparameter} & \textbf{Value} \\
\midrule
Backbone LLM & \texttt{gpt-oss-120b} \\
Prompt template & Same as \ourmethod \\
Decoding temperature & \(0.1\) \\
Routing granularity & Reasoning step \\
Routing signal & Logit-based confidence \\
Confidence estimator & Maximum logit confidence \\
Calibration model & Two-component Gaussian mixture model \\
External router & No \\
Routing threshold search interval & \(0.10\) \\
Routing threshold selection & Validation-selected \\
Fallback strategy & Route to high-confidence branch \\
Training budget & \(40\)--\(80\) queries \\
\bottomrule
\end{tabular}}
\caption{Key hyperparameter settings of STEER.}
\label{tab:steer_hparams}
\end{table}

\begin{table}[!htbp]
\centering
\small
\resizebox{0.65\columnwidth}{!}{
\begin{tabular}{lc}
\toprule
\textbf{Hyperparameter} & \textbf{Value} \\
\midrule
Backbone LLM & \texttt{gpt-oss-120b} \\
Prompt template & Same as \ourmethod \\
Decoding temperature & \(0.1\) \\
Initial topology & Fully connected graph \\
Pruning granularity & Communication edge \\
Initial edge scoring & Semantic compatibility \\
Adaptive edge scoring & Historical contribution \\
Clustering strategy & 0-extension clustering \\
Pruning schedule & Progressive pruning \\
Greedy top-k pruning & No \\
Training budget & \(40\)--\(80\) queries \\
Final topology selection & Validation-selected \\
\bottomrule
\end{tabular}}
\caption{Key hyperparameter settings of SafeSieve.}
\label{tab:safesieve_hparams}
\end{table}

\begin{table}[!htbp]
\centering
\small
\resizebox{0.65\columnwidth}{!}{
\begin{tabular}{lc}
\toprule
\textbf{Hyperparameter} & \textbf{Value} \\
\midrule
Backbone LLM & \texttt{gpt-oss-120b} \\
Prompt template & Same as \ourmethod \\
Decoding temperature & \(0.1\) \\
Topology generation paradigm & Autoregressive graph generation \\
Graph initialization & Empty graph \\
Node generation strategy & Role selection from agent pool \\
Edge generation strategy & Sequential edge prediction \\
Node encoder & \texttt{all-MiniLM-L6-v2} \\
Node embedding dimension & \(384\) \\
Communication rounds & \(3\) \\
Training strategy & Two-stage curriculum learning \\
Training objective & Negative log-likelihood \\
Node-edge loss weight & \(0.20\) \\
Training budget & \(40\)--\(80\) queries \\
Final aggregation & Summarizer agent \\
\bottomrule
\end{tabular}}
\caption{Key hyperparameter settings of ARG-Designer.}
\label{tab:argdesigner_hparams}
\end{table}

\end{document}